\documentclass{article}

\usepackage{PRIMEarxiv}

\usepackage[utf8]{inputenc} 
\usepackage[T1]{fontenc}    
\usepackage{hyperref}       

\usepackage{graphicx}
\usepackage{booktabs}

\usepackage{mathtools}
\usepackage{multirow}
\usepackage{graphicx}
\usepackage{lscape}

\usepackage{graphicx}
\usepackage{subcaption}

\usepackage[table,xcdraw]{xcolor}

\usepackage[capitalize,noabbrev]{cleveref}

\title{TS40K: a 3D Point Cloud Dataset of Rural Terrain and Electrical Transmission System
}

\author{
Diogo Lavado \\
NOVA School of Science and Technology, Lisbon \\
University of Milan, Milan\\
\texttt{d.lavado@campus.fct.unl.pt} 
\And
Cláudia Soares \\
NOVA School of Science and Technology \\
Lisbon\\
\texttt{claudia.soares@fct.unl.pt} 
\And
Alessandra Micheletti \\
University of Milan,\\
Milan\\
\texttt{alessandra.micheletti@unimi.it} \\
\And
Ricardo Santos \\
EDP - Labelec,\\
Lisbon\\
\texttt{ricardovieira.santos@edp.com}
\And
André Coelho \\
EDP - Labelec,\\
Lisbon\\
\texttt{andre.coelho@edp.com}
\And
João Santos \\
CNET - CENTRE NEW ENERGY,\\
Lisbon\\
\texttt{joao.passagemsantos@edp.pt}
}

\begin{document}
\maketitle

\begin{abstract}
Research on supervised learning algorithms in 3D scene understanding has risen in prominence and witness great increases in performance across several datasets. The leading force of this research is the problem of autonomous driving followed by indoor scene segmentation.
However, openly available 3D data on these tasks mainly focuses on urban scenarios. In this paper, we propose TS40K, a 3D point cloud dataset that encompasses more than 40,000 Km on electrical transmission systems situated in European rural terrain.
This is not only a novel problem for the research community that can aid in the high-risk mission of power-grid inspection, but it also offers 3D point clouds with distinct characteristics from those in self-driving and indoor 3D data, such as high point-density and no occlusion.
In our dataset, each 3D point is labeled with 1 out of 22 annotated classes.
We evaluate the performance of state-of-the-art methods on our dataset concerning 3D semantic segmentation and 3D object detection.
Finally, we provide a comprehensive analysis of the results along with key challenges such as using labels that were not originally intended for learning tasks.
\end{abstract}

\section{Introduction}\label{sec:intro}

3D scene understanding stands as a pivotal challenge in computer vision research, with machines striving to accurately recognize and categorize our intricate three-dimensional world. Key applications, such as autonomous driving~\cite{mao20223d} and indoor scene segmentation~\cite{armeni2017joint,dai2017scannet}, have driven groundbreaking breakthroughs.
However, the current landscape predominantly revolves around the most prominent publicly available datasets, centering on self-driving scenarios and indoor environments. 
Notably, models like JS3C-Net~\cite{yan2021sparse} and RangeFormer~\cite{kong2023rethinking} tailor their architectures to respond to the challenging scenarios akin to a vehicle's point of view in urban settings. While this focus has propelled advancements, it also subtly narrows the scope of exploration, limiting the diversity of benchmarks.
%
\begin{figure}[]
\centering
\begin{subfigure}[b]{1.0\columnwidth}
  \includegraphics[width=\linewidth]{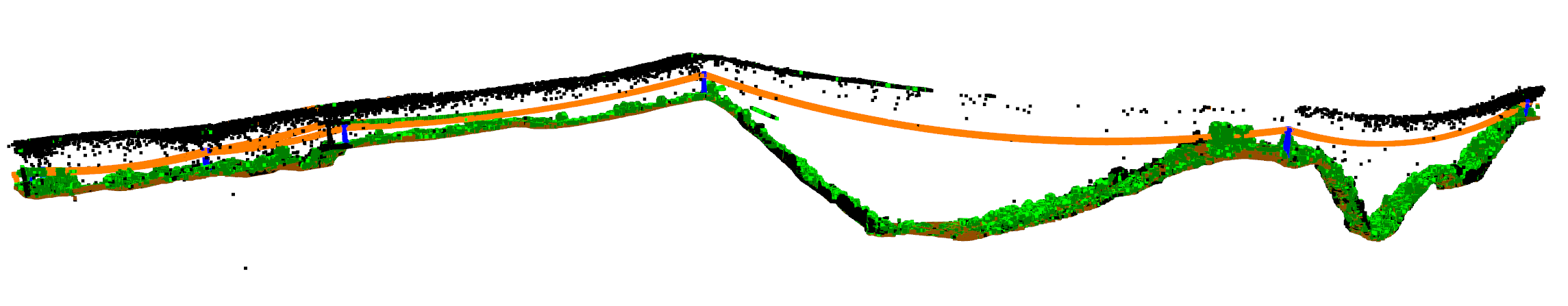}
  \caption{raw TS40K sample}
  \label{fig:teaser-raw-sample}
\end{subfigure}
\hfill
\begin{subfigure}[b]{0.46\columnwidth}
  \includegraphics[width=\linewidth]{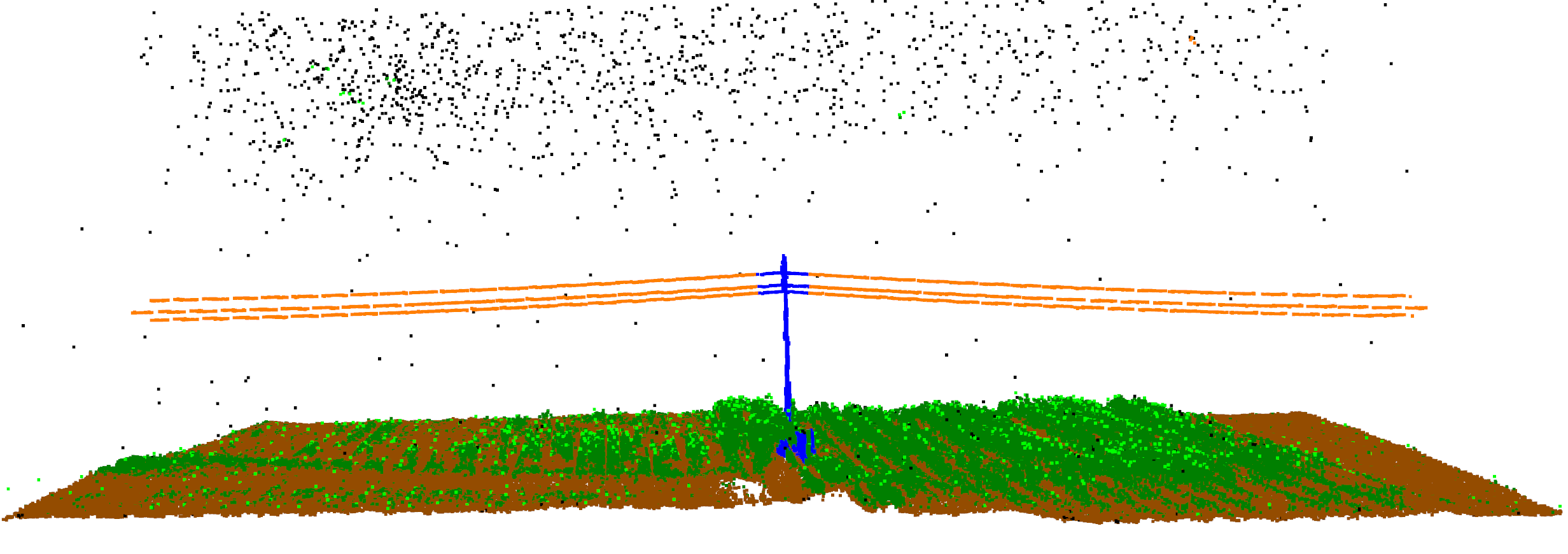}
  \caption{Tower-radius}
  \label{fig:teaser-tower-radius}
\end{subfigure}
\hfill
\begin{subfigure}[b]{0.42\columnwidth}
  \includegraphics[width=\linewidth]{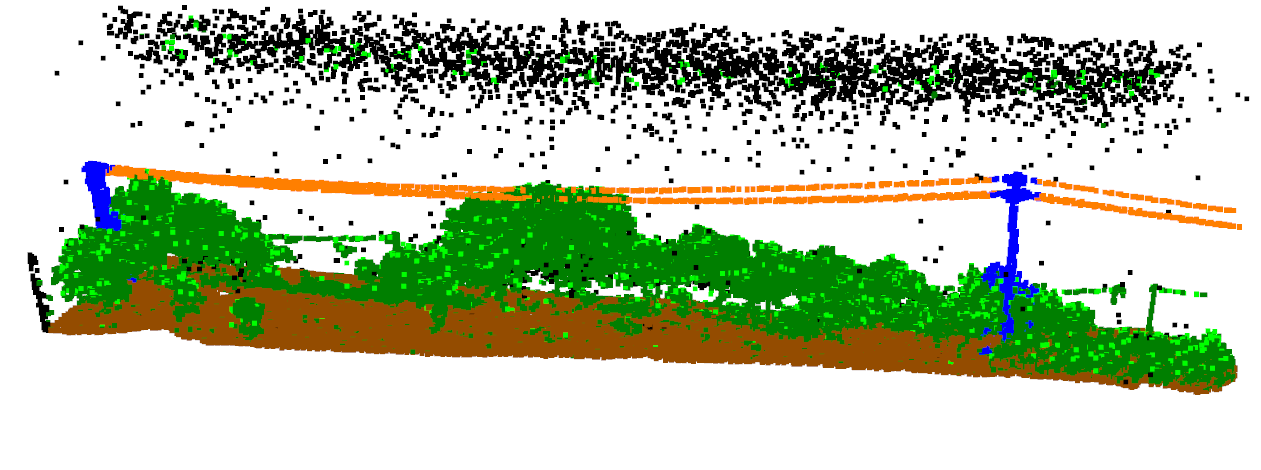}
  \caption{Power-line}
  \label{fig:teaser-power-line}
\end{subfigure}
\hfill
\begin{subfigure}[b]{0.35\columnwidth}
  \includegraphics[width=\linewidth]{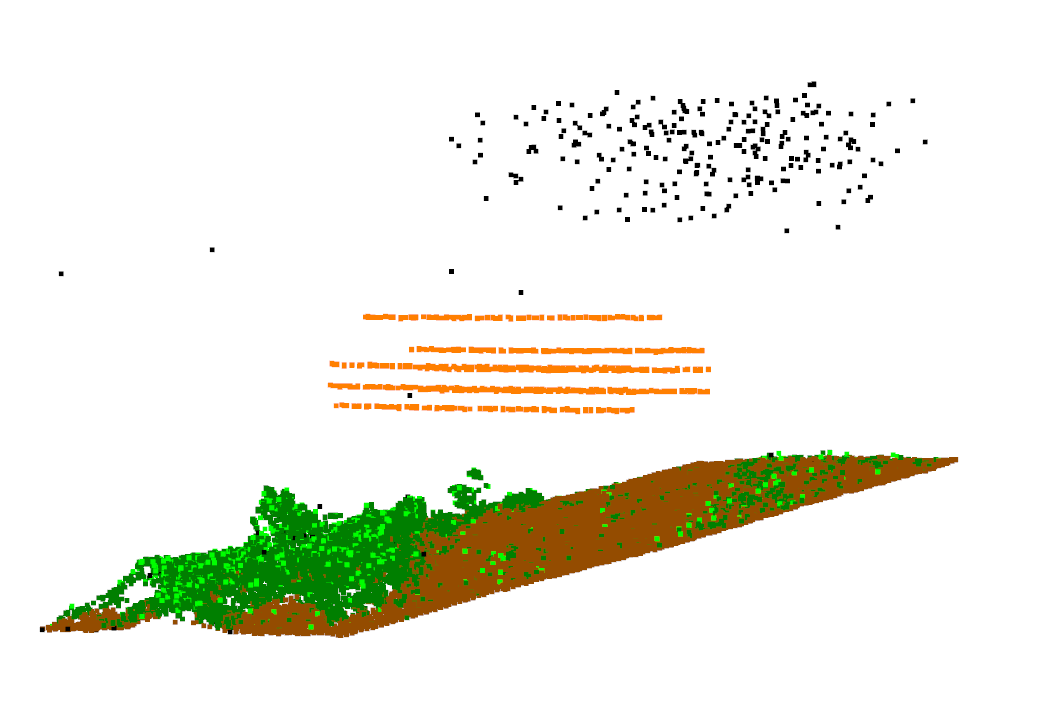}
  \caption{No-tower}
  \label{fig:teaser-no-ts}
\end{subfigure}

\includegraphics[width=1\columnwidth]{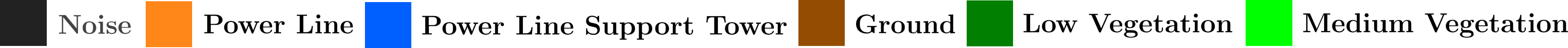}
\begin{subfigure}[b]{0.48\columnwidth}
  \includegraphics[width=\linewidth]{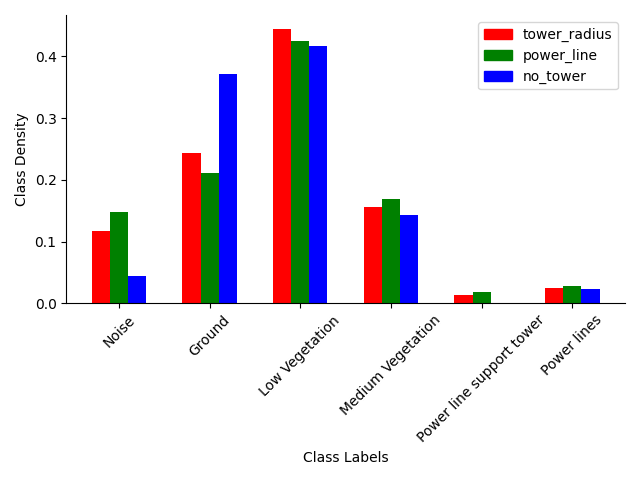}
  \caption{Sample type density for each label}
  \label{fig:density1}
\end{subfigure}
\hfill
\begin{subfigure}[b]{0.48\columnwidth}
  \includegraphics[width=\linewidth]{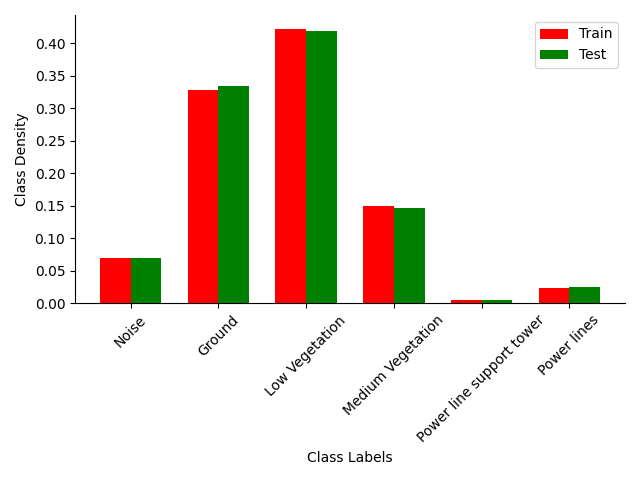}
  \caption{Overall train/test class density}
  \label{fig:density2}
\end{subfigure}
\caption{The TS40K dataset is derived from raw 3D scans illustrated in Figure~\ref{fig:teaser-raw-sample} and processed into three different sample types: \textit{(1) Tower-radius} focuses on the towers that support power lines and its environment (Fig.~\ref{fig:teaser-tower-radius}). Conversely, \textit{(2) Power-line} samples have power lines as their main focus in the 3D scenes (Fig.~\ref{fig:teaser-power-line}). Lastly, \textit{(3) No-tower} samples represent rural terrain where the transmission system is located, excluding supporting towers but potentially including power lines (Fig.~\ref{fig:teaser-no-ts}).
In Figures~\ref{fig:density1} and~\ref{fig:density2}, we showcase the semantic class densities of the TS40K dataset. Figure~\ref{fig:density1} illustrates the class density for each of the aforementioned sample types. In turn, Figure~\ref{fig:density2} shows the overall class density in the TS40K train and test sets.}
\label{fig:teaser}
\end{figure}

To invigorate the field, we recognize the need for diverse benchmarks that go beyond urban environments and conventional object representations. Herein lies our main contribution—the introduction of \textbf{TS40K}, an expansive outdoor 3D point cloud dataset that spans over 40,000 kilometers of European electrical transmission system in rural terrain.
%
TS40K boosts advancements in both 3D semantic segmentation, featuring per-point annotations across five semantic classes, and 3D object detection, comprehensively covering crucial elements such as the power grid.
%
3D scene understanding within electrical transmission systems enhances inspection efficacy, safeguarding against power outages, grid damages, and, most importantly, forest fires.  
Electrical companies now opt for cost-effective drone-based inspections over traditional on-site methods. Unmanned aerial vehicles (UAVs) equipped with LiDAR technology 
capture detailed power grid scans, which undergo meticulous processing by maintenance personnel.  
This manual process is time-consuming, involving the annotation of each 3D point with inspection-critical labels, which facilitates an in-depth examination of the potential risks of contact between the transmission system and its surrounding environment. 
Leveraging 3D scene understanding models can lead to faster and more efficient power grid maintenance and forest fire prevention, addressing the challenges and time constraints of manual inspections.

Besides the novelty in 3D data that this benchmark provides, there are key challenges that our dataset TS40K raises:
\textbf{(1) Different 3D data properties.} 
TS40K poses distinctive challenges that stem from its unique properties, the use of UAVs to capture power grid environments resulting in a 3D point cloud featuring characteristics not typically found in self-driving benchmarks— high point-density, absence of object occlusion, and homogeneous point-density in objects. Understanding how current state-of-the-art methods adapt to these conditions is a compelling avenue for exploration.
\textbf{(2) Inspection-based annotations.}
Our dataset brings inspection-based annotations, crafted not for AI models but by maintenance personnel. The mapping of these labels to semantically significant classes introduces challenges such as noise and mislabeled points, offering a realistic evaluation of methods on real-world data.
\textbf{(3) Extreme Class Imbalance.} 
TS40K further accentuates the challenge of class imbalance inherent in rural scenarios, where the ground and low-medium vegetation dominate, while classes related to the transmission system are inherently underrepresented  (around 1.4\% of the overall dataset). 

Building upon TS40K, we discern several crucial challenges and limitations and empirically explore them in Section~\ref{sec:challenges}. 
Particularly, discuss the methodology employed in building the TS40K dataset, emphasizing the preservation of the security and safety of European power grid systems while ensuring its public availability for research. 
Second, we investigate the impact of the extreme class imbalance in TS40K and how this imbalance influences neural networks in our experiments.
Lastly, we evaluate whether relying solely on geometry is sufficient to achieve satisfactory performance in the TS40K dataset.

In summary, our main contributions include:
\begin{enumerate}
    \item \textbf{Novel Dataset:} We present the first publicly available 3D point cloud dataset focusing on electrical transmission systems in rural areas.
    \item \textbf{Method Evaluation:} We provide a comprehensive evaluation of state-of-the-art methods for 3D semantic segmentation and 3D object detection applied to point clouds.
    \item \textbf{Challenges and Future Directions:} We highlight the challenges encountered and outline future directions in 3D scene understanding, using our TS40K dataset as a catalyst for research exploration.
\end{enumerate}

In essence, our work not only introduces a groundbreaking dataset but also extends an invitation to researchers to explore, innovate, and overcome challenges in a broader, more diverse landscape. 

\section{Related Work}

\subsection{Contemporary 3D Datasets}

The progress in 3D scene understanding techniques is largely substantiated by high-quality, large-scale and densely annotated datasets~\cite{torralba2011unbiased}.
3D benchmarks currently available can be broadly categorized into three groups:
\textbf{(1) Outdoor urban datasets.} Most of these datasets focus on self-driving applications and complex urban structures, namely, the KITTI benchmark~\cite{KITTI1,KITTI2,SemKITTI}, KAIST~\cite{KAIST}, Lyft dataset~\cite{lyft2019}, Cityscapes 3D~\cite{gahlert2020cityscapes}, nuScenes~\cite{caesar2020nuscenes}, Waymo dataset~\cite{WAYMO}, SensatUrban~\cite{SensatUrban} and the ONCE dataset~\cite{mao2021one}.
\textbf{(2) Indoor 3D scans.} 
These benchmarks include heavily detailed indoor environments, such as NYU3D~\cite{silberman2012indoor},
SUN RGB-D~\cite{song2015sun}, 
SceneNN~\cite{hua2016scenenn},
S3DIS~\cite{armeni2017joint}
and ScanNet~\cite{dai2017scannet}.
\textbf{(3) 3D object representations.}
These datasets include every-day objects for classification or part-segmentation, for instance, MeshSeg~\cite{chen2009benchmark}, 
ShapeNet~\cite{chang2015shapenet},
PartNet~\cite{mo2019partnet},
and ScanObjectNN~\cite{uy2019revisiting}.
Conversely to these benchmarks, there are few publicly available datasets that focus on non-urban areas and none that follow power grid systems.
For example, 3D datasets that cover rural or forest terrain are Forest3D~\cite{trochta20173d}, GTASynth~\cite{curnis2022gtasynth} and
NEON~\cite{marconi2019data}.
Forest3D~\cite{trochta20173d} focuses solely on 3D representations of trees and in their instance segmentation. GTASynth~\cite{curnis2022gtasynth} is a synthetic dataset with of non-urban environments with 3D point cloud properties similar to self-driving datasets, specifically low point-density, object occlusion and captured from a vehicle point of view. Lastly, NEON~\cite{marconi2019data} captures airborne LiDAR tree data from a birds-eye view perspective to predict tree crown dimensions.
In contrast to these datasets, the TS40K dataset focuses on electrical power grid systems and their environment, which include low and medium vegetation, trees, and highly irregular terrains. 
Additionally, TS40K  is composed of medium and large point clouds that feature high point-density, no object occlusion, and homogeneous object density. However, they also present class imbalance in power-grid elements. Conversely, urban datasets usually contain sparse point clouds with occluded objects, limiting performance, while indoor and 3D object datasets tend to focus on small point clouds within closed environments.
A comprehensive comparison between prominent publicly available benchmarks and TS40K can be found in Table~\ref{tab:datasets}.

\begin{table}[ht]
\centering
\caption{Key attributes of various 3D datasets prominent in 3D scene understanding. Entries include the dataset's publication year, viewing perspective, realism (real or simulated), total point count, number of semantic classes and the number of annotated classes in brackets, bounding box cardinality, and inclusion of RGB information.
Forest datasets were excluded from this overview due to the absence of crucial information, such as the number of points or classes and the number of bounding boxes.}
\resizebox{\textwidth}{!}{%
\begin{tabular}{l|c|c|c|c|c|c|c}
\hline
\multicolumn{1}{c|}{3D Dataset}                          & Year & View           & Real/Simulated & \#Points                 & \#Classes & \#Bounding Boxes         & RGB \\ \hline
KITTI~\cite{KITTI1, KITTI2}            & 2012 & single vehicle & Real           & 1799M                    & 3 (8)     & 200K                     & No  \\
KAIST~\cite{KAIST}                     & 2018 & single vehicle & Real           & -                        & 3 (3)     & 103K                     & Yes \\
SemanticKITTI\cite{SemKITTI}           & 2019 & single vehicle & Real           & 4549M                    & 25 (28)   & -                        & No  \\
Lyft~\cite{lyft2019}                   & 2019 & single vehicle & Real           & -                        & 9 (9)     & 1.3M                     & No  \\
Waymo Open~\cite{WAYMO}                & 2019 & single vehicle & Real           & -                        & 4 (4)     & 12M                      & No  \\
SensatUrban~\cite{SensatUrban}    & 2020 & UAV            & Real           & 2847M                    & 13 (31)   & -                        & Yes \\ \hline
ScanNet~\cite{dai2017scannet}          & 2017 & RGB-D          & Real           & 242M                     & 20 (20)   & -                        & Yes \\
S3DIS~\cite{armeni2017joint}           & 2017 & RGB-D          & Real           & 273M                     & 13 (13)   & -                        & Yes \\ \hline
ShapeNet~\cite{chang2015shapenet} & 2015 & -              & Simulated      & -                        & 55 (55)   & -                        & No  \\
PartNet~\cite{mo2019partnet}      & 2019 & -              & Simulated      & -                        & 24 (24)   & -                        & No  \\ \hline
\textbf{TS40K (Ours)}                                   & 2024 & UAV            & Real           & 2595M & 5 (22)    & 36K & No  \\ \hline
\end{tabular}%
}
\label{tab:datasets}
\end{table}

\subsection{3D Semantic Segmentation}

Semantic segmentation, operating at the scene level, seeks to partition a 3D point cloud into subsets based on the semantic meanings attributed to individual points.
Semantic segmentation methodologies can be broadly categorized into four paradigms~\cite{guo2020deep}: 
\textbf{(1) projection-based.}~\cite{su2015multi,lawin2017deep,yang2019learning,lyu2020learning} These methods employ established 2D CNN frameworks to learn 3D semantics. However, projecting point clouds onto 2D images introduces the risk of losing critical geometric information.
\textbf{(2) discretization-based.}~\cite{choy20194d,zhou2018voxelnet,le2018pointgrid,meng2019vv,zhang2020polarnet}
These models leverage 3D CNN architectures. While effective, these techniques often encounter scalability challenges  due to significant computational and memory requirements.
\textbf{(3) point-based.}~\cite{qi2017pointnet,qi2017pointnet++,li2018pointcnn,thomas2019kpconv,hu2020randla,kong2023rethinking,lai2023spherical,wu2023ptv3}
These techniques adopt the use Multi-Layer Perceptrons (MLPs) and transformers to learn individual point-wise semantics.
In contrast to voxel and projection-based methods, point-based architectures preserve the semantics for each individual 3D point and achieve state-of-the-art performance in most prominent datasets.
Lastly, \textbf{(4) hybrid methods.}~\cite{dai20183dmv,jaritz2019multi,tang2020searching,hou2022point}
These methods take advantage of the three techniques described above and fuse the feature extraction knowledge from each channel to achieve better domain understanding of 3D scenes.

In the realm of 3D semantic segmentation, new methods emerge regularly, challenging established benchmarks, particularly those tailored for autonomous driving. These benchmarks dominate the field, shaping research priorities towards improving performance in such scenarios.
To diversify the evaluation landscape, we introduce an alternative 3D dataset that deviates from the typical self-driving focus.
Our dataset addresses the specific challenges of power-grid inspection, a critical task for preventing power outages and forest fires.

\subsection{3D Object Detection}

3D object detection aims to predict bounding boxes of 3D objects in 3D scenes. A general formula of 3D object detection can be represented as
\begin{equation}
    B = f_{det}(\mathcal{I}), \label{eq:object_detection}
\end{equation}
where $B = \{B_1, \ldots, B_N\}$ is a set of $N$ 3D objects in a scene, $f_{det}$ is a 3D object detection model, and $\mathcal{I}$ is one or more sensory inputs. A 3D object $B_i$ is represented as a 3D cuboid, including its parameters as follows:

\begin{equation}
    B_i = [x_c, y_c, z_c, l, w, h, \theta, \text{class}], \label{eq:cuboid_parameters}
\end{equation}
where $(x_c, y_c, z_c)$ is the 3D center coordinate of the cuboid, $l$, $w$, $h$ represent its length, width, and height, respectively. $\theta$ is the heading angle of the cuboid on the ground plane, and \text{class} denotes the category of the 3D object~\cite{mao20223d}.

3D object detection methodologies can be broadly categorized in two groups~\cite{guo2020deep}:
\textbf{(1) Region proposal-based methods.}~\cite{shi2019pointrcnn,yang2018ipod,zarzar2019pointrgcn,zhou2018voxelnet,shi2023pv}
These techniques initially generate multiple potential regions, often referred to as proposals, that encompass objects. Subsequently, they extract features specific to each region to ascertain the category label associated with each proposal.
\textbf{(2) Single shot methods.}~\cite{li20173d,engelcke2017vote3deep,yang2018hdnet,xu2022behind,zhang2023glenet}
These approaches directly forecast class probabilities and perform 3D bounding box regression using a single-stage network. Dispensing the need for a region proposal stage and subsequent post-processing, they show accelerated processing speed.
As indicated in Table~\ref{tab:datasets}, 3D bounding box annotations for objects within 3D scenes are predominantly available in autonomous driving benchmarks. Our TS40K dataset contributes to broadening the scope of 3D object detection benchmarks by introducing a focus on power grid elements in a rural setting. Notably, we offer 3D bounding boxes for power lines, their supporting towers and medium vegetation. This inclusion enhances the ability of inspectors to assess the risk of grid contact with its surroundings.

\section{The TS40K Dataset}

\begin{figure}[ht]
\centering
\begin{subfigure}[b]{1.0\columnwidth}
  \includegraphics[width=\linewidth]{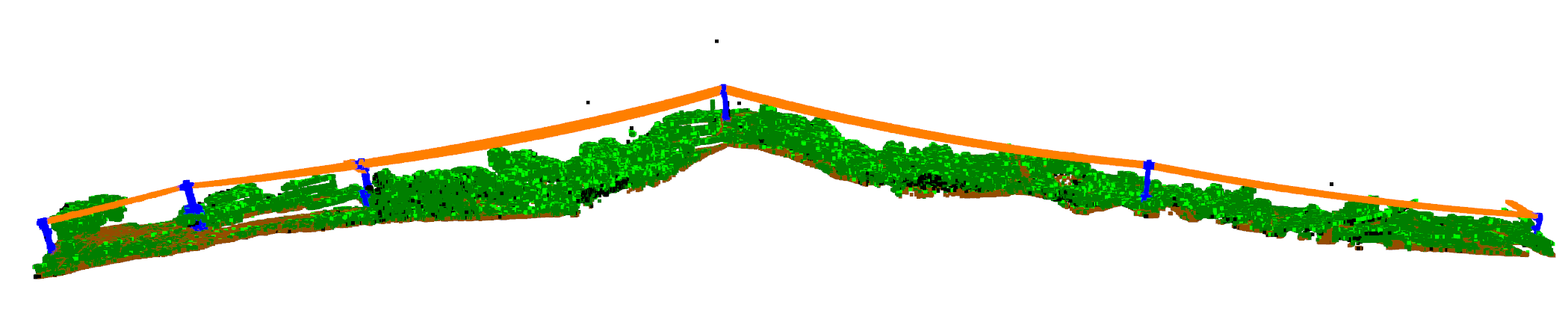}
\end{subfigure}
\hfill
\begin{subfigure}[b]{0.65\columnwidth}
  \includegraphics[width=\linewidth]{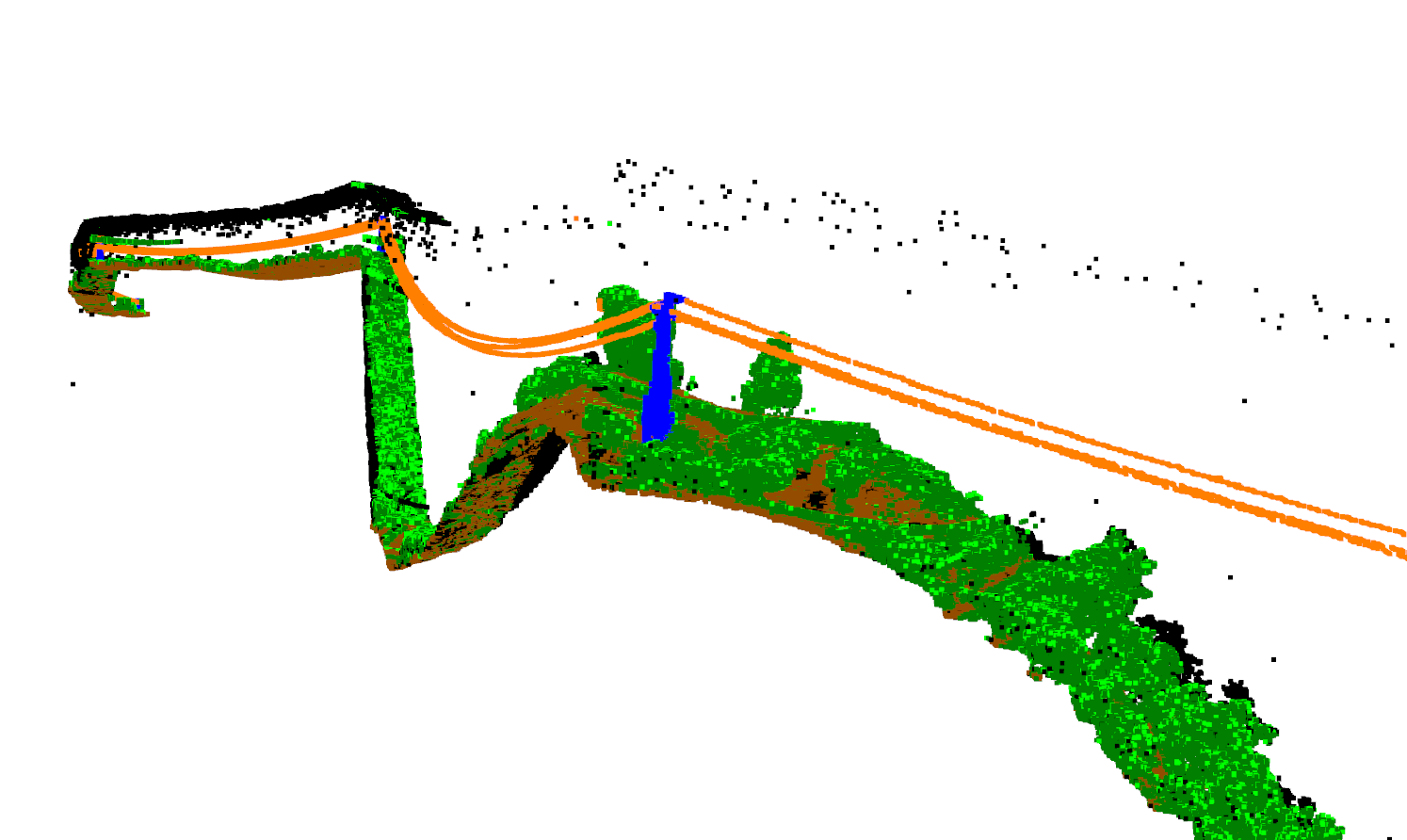}
\end{subfigure}

\includegraphics[width=1\columnwidth]{legend_classes.png}
\\
\caption{Examples of TS40K raw 3D point clouds. Different semantic classes are labeled with different colors.}
\label{fig:ts40k}
\end{figure}

\subsection{3D Point Cloud Collection}

\begin{table}[ht]
\centering
\caption{Available classes in the TS40K dataset and their distribution. Ground and road surfaces constitute the majority of the dataset (63\%), whereas the power grid, i.e., the power lines and supporting towers only constitute 1.43\% of the 3D points.}
\resizebox{0.7\textwidth}{!}{%
\begin{tabular}{lll|lll}
\hline
Label &
  Class &
  Density(\%) &
  Label &
  Class &
  Density(\%) \\ \hline
0  & Created          & 0               & 11 & Road surface                    & \textbf{44.752} \\
1  & Unclassified     & 0.571           & 12 & Overlap points                  & 0.529           \\
2  & Ground           & \textbf{23.403} & 13 & Medium Reliability              & 0               \\
3  & Low vegetation   & \textbf{18.758}           & 14 & Low Reliability                 & 0               \\
4 &
  Medium vegetation &
  0.241 &
  15 &
  Power line support tower &
  {\color[HTML]{CB0000} \textbf{0.519}} \\
5  & Natural obstacle & 1.069           & 16 & Main power line                 & 
 {\color[HTML]{CB0000} \textbf{0.907}}           \\
6  & Human structures & 0               & 17 & Other power line                & {\color[HTML]{CB0000} \textbf{0.002}}       \\
7  & Low point        & 0.362         & 18 & Fiber optic cable               & 0               \\
8  & Model key points & 0             & 19 & Not rated object to be consider & 8.205           \\
9  & Water            & 0             & 20 & Not rated object to be ignored  & 0               \\
10 & Rail             & 0.681         & 21 & Incidents                       & 0               \\ \hline
\end{tabular}%
}
\label{tab:ts40k_labels}
\end{table}

The TS40K dataset utilizes 3D point cloud data obtained from unmanned aerial vehicles (UAVs) conducting scans of European electrical transmission systems. 
Notably, the use of UAVs results in capturing data from a birds-eye view perspective, this leads to good data characteristics for learning models, such as high point-density, absence of object occlusion, and homogeneous object density.
Our dataset comprises over 500 processed and annotated scans by maintenance personnel and encompasses 40,000 km of diverse land strips dedicated to the transmission system (Figure~\ref{fig:ts40k}).
Each 3D point is labeled from a set of 22 classes detailed in Table~\ref{tab:ts40k_labels}.

\subsection{Point Cloud Annotations by Maintenance Personnel}

\begin{figure}[t]
\centering
\begin{subfigure}[b]{0.30\columnwidth}
  \includegraphics[width=\linewidth]{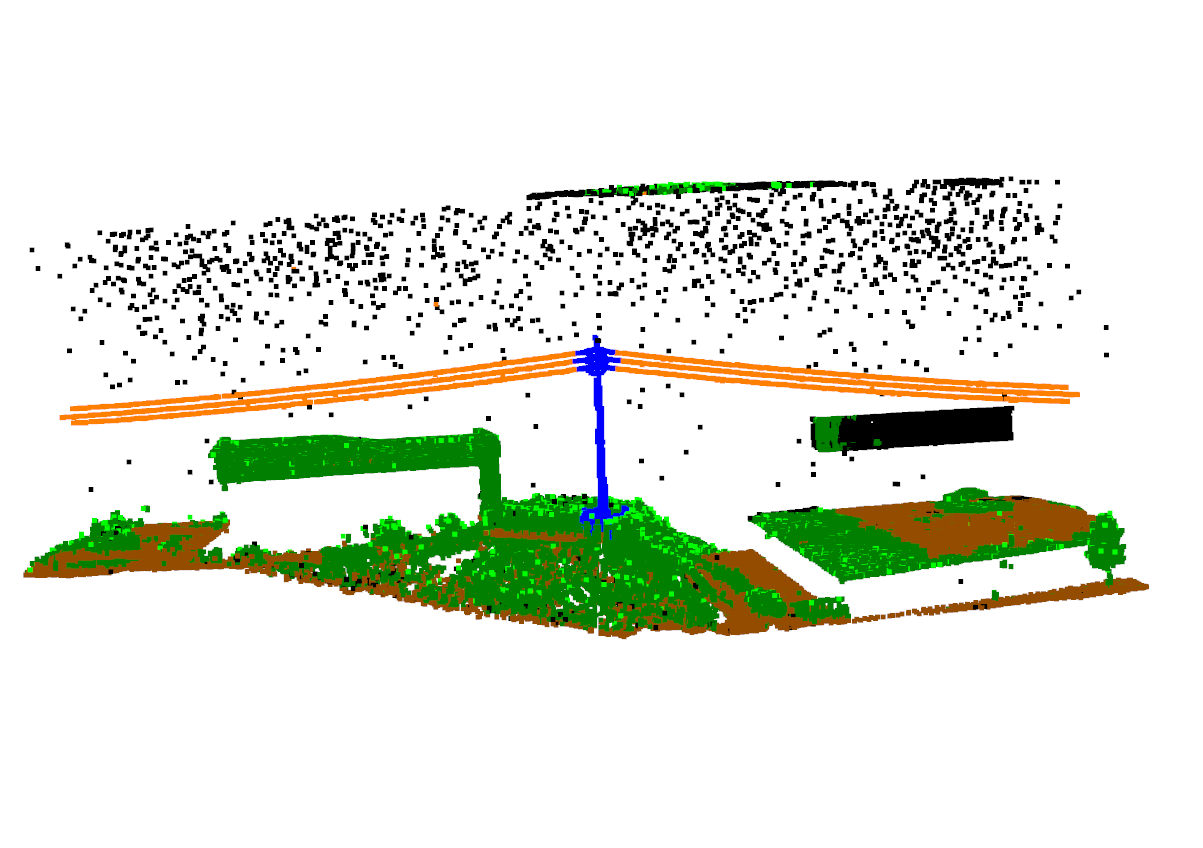}
\end{subfigure}
\hfill
\begin{subfigure}[b]{0.40\columnwidth}
  \includegraphics[width=\linewidth]{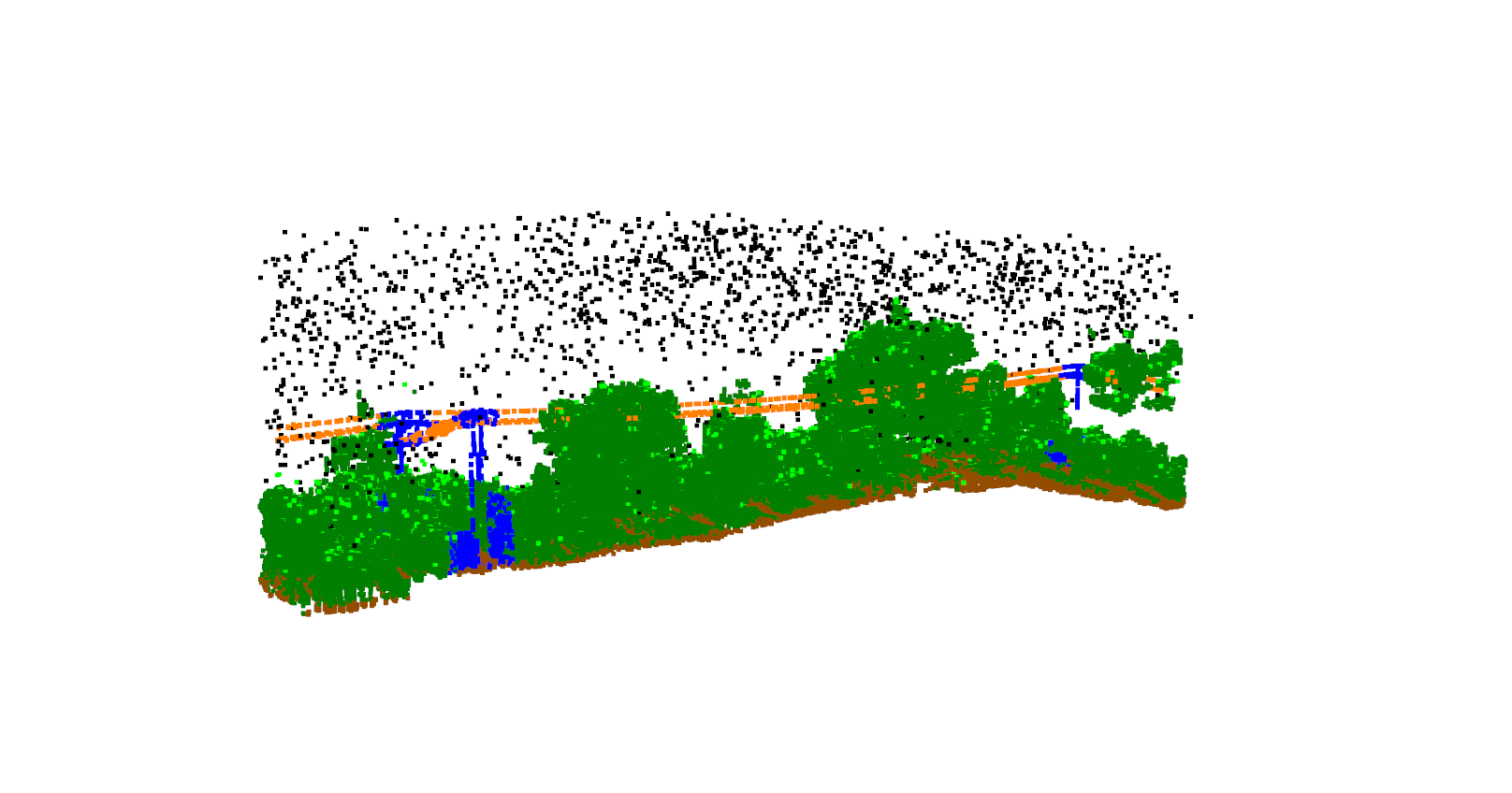}
\end{subfigure}
\hfill
\begin{subfigure}[b]{0.28\columnwidth}
  \includegraphics[width=\linewidth]{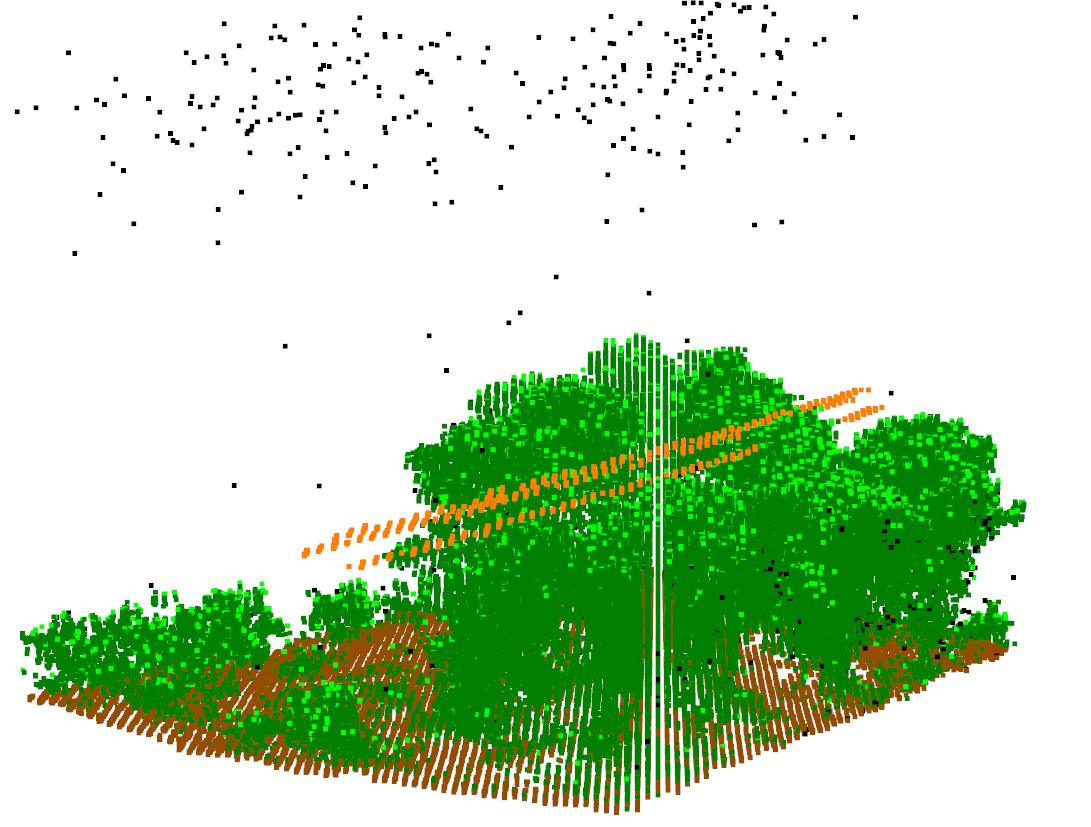}
\end{subfigure}
\\
\begin{subfigure}[b]{0.35\columnwidth}
  \includegraphics[width=\linewidth]{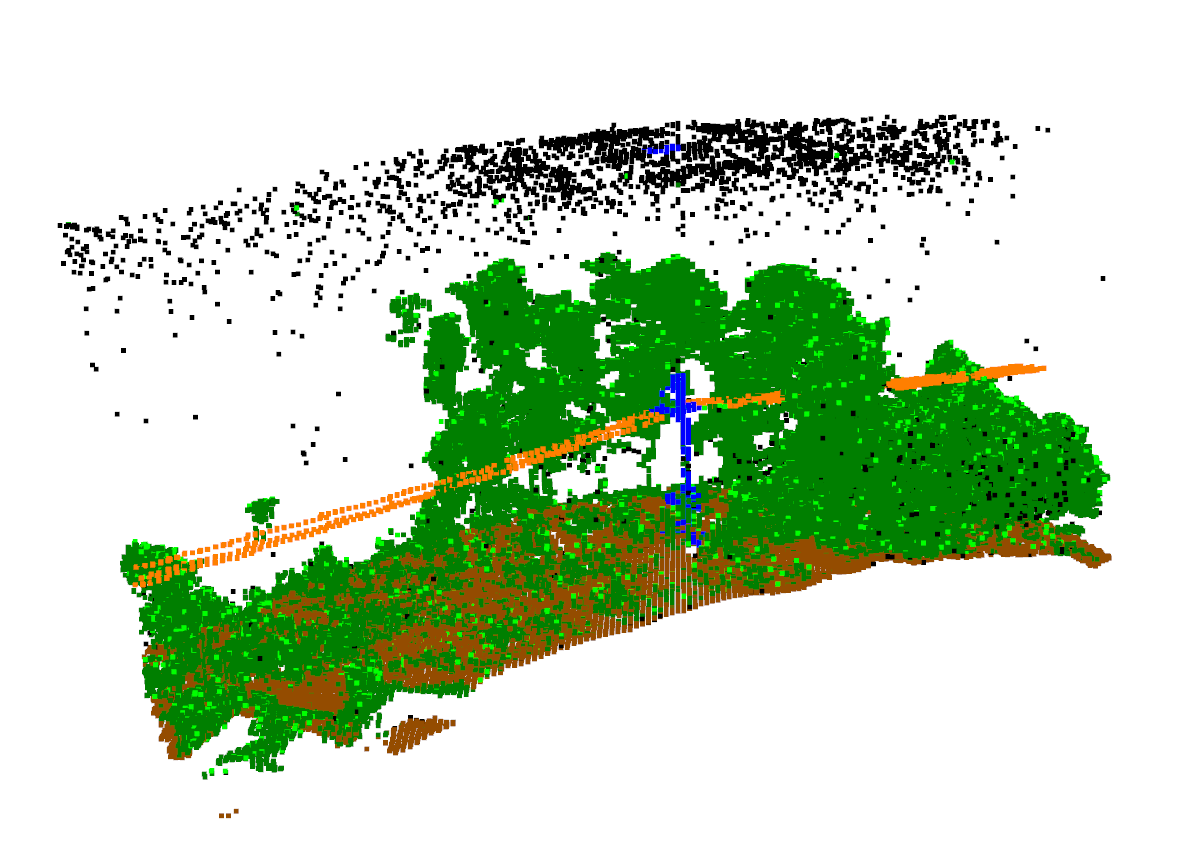}
  \caption{Tower-radius}
  \label{fig:tower-radius}
\end{subfigure}
\hfill
\begin{subfigure}[b]{0.35\columnwidth}
  \includegraphics[width=\linewidth]{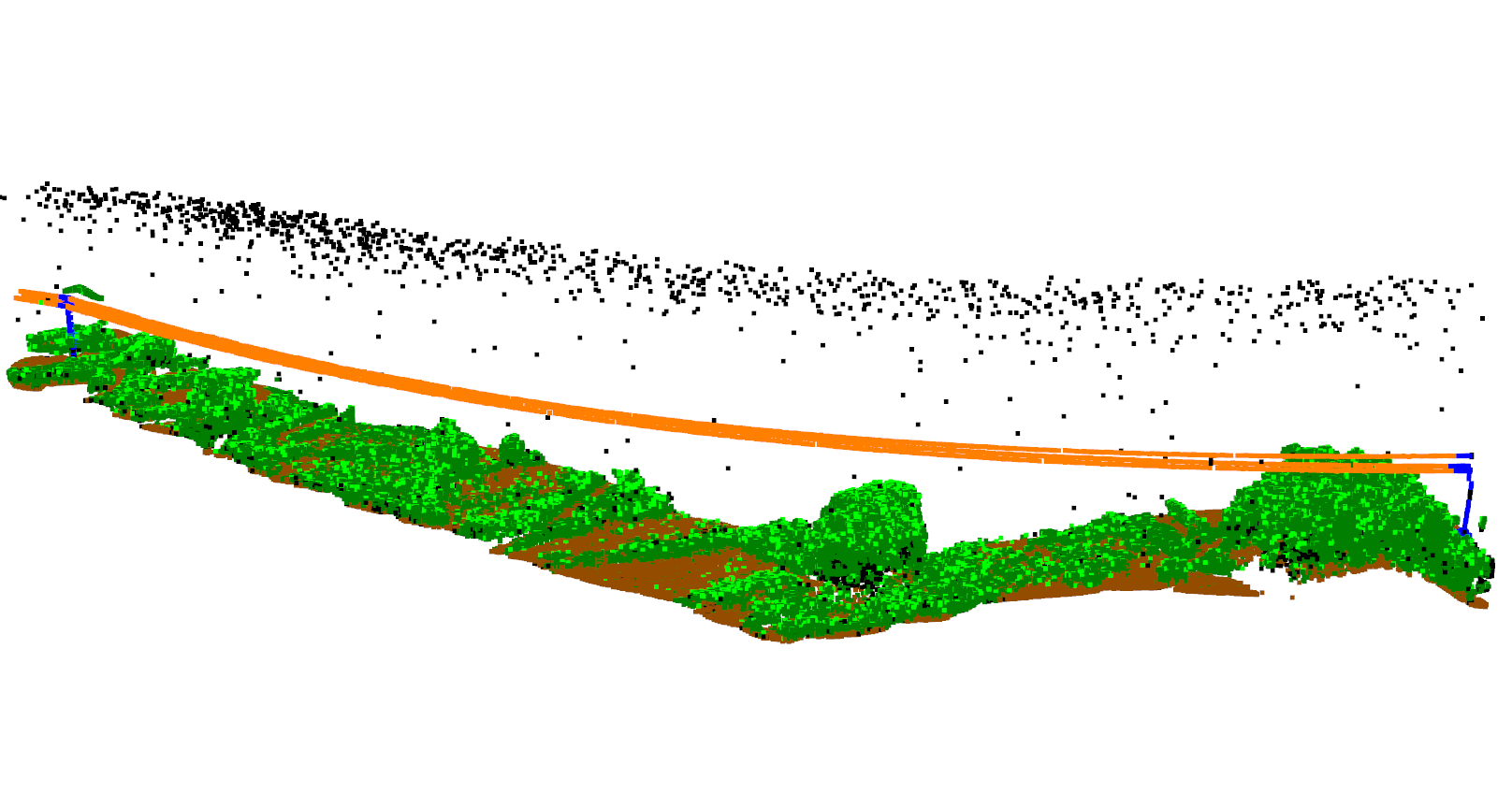}
  \caption{Power-line}
  \label{fig:power-line}
\end{subfigure}
\hfill
\begin{subfigure}[b]{0.28\columnwidth}
  \includegraphics[width=\linewidth]{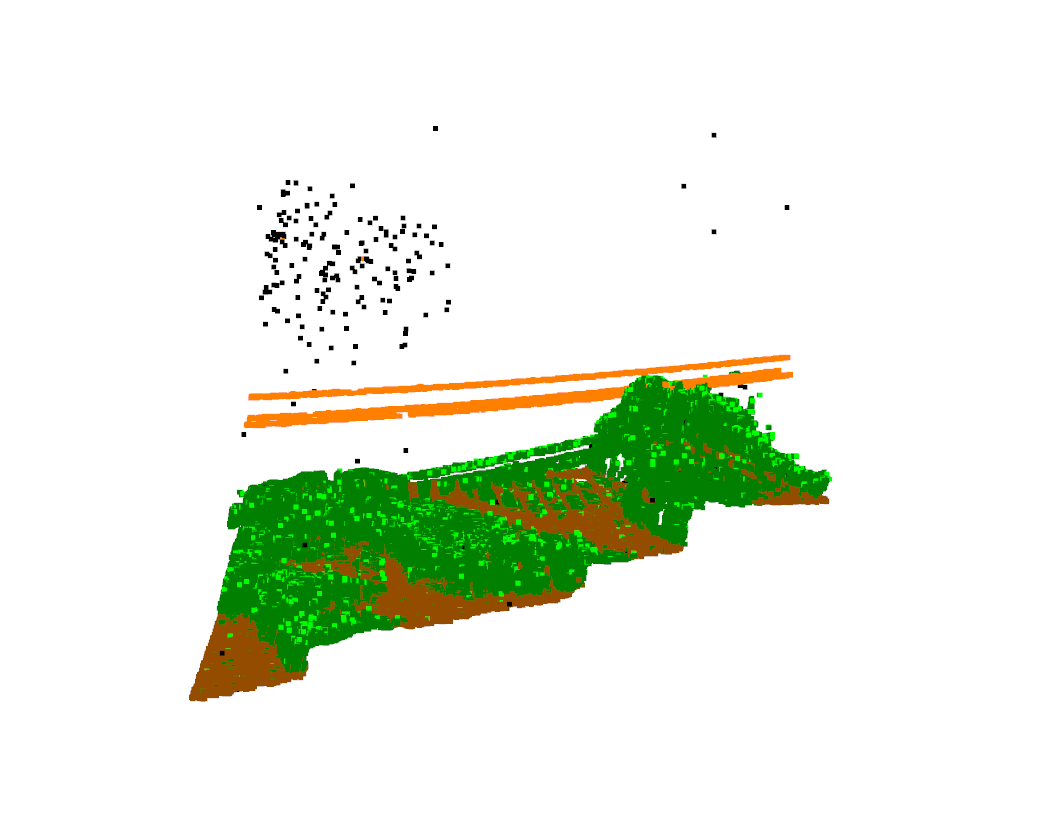}
  \caption{No-tower}
  \label{fig:no-ts}
\end{subfigure}

\includegraphics[width=1\columnwidth]{legend_classes.png}
\\
\caption{The raw TS40K land strips are partitioned into three sample types: \textit{Tower-radius}: Encompasses areas around power-line supporting towers (Fig.~\ref{fig:tower-radius}); \textit{Power-line}: Focuses on power lines between towers (Fig.~\ref{fig:power-line}); and \textit{No-tower}: Represents rural areas without towers but potentially including power lines(Fig.~\ref{fig:no-ts}). This categorization ensures safety and addresses data imbalance.}
\label{fig:ts40k-sample-types}
\end{figure}

\begin{table}[ht]
\centering
\caption{Distribution of semantic classes in the TS40K dataset across the three sample types: tower-radius, power-line, and no-tower.}
\resizebox{0.45\textwidth}{!}{%
\label{tab:ts40k_sem_labels}
\begin{tabular}{cll}
\hline
\multicolumn{1}{l}{Label} & Semantic Class            & Density (\%) \\ \hline
0                         & Noise                     & 1.348        \\
1                         & Ground                    & 55.281       \\
2                         & Low Vegetation            & 35.520       \\
3                         & Medium Vegetation         & 6.647        \\
4                         & Power Line Support Tower  & 0.431        \\
5                         & Power Line                & 0.771        \\ \hline
\end{tabular}
}
\end{table}
The annotations made by maintenance personnel originally aimed to expedite power grid inspections and not to train machine learning models. Thus, we mapped the provided maintenance classes to semantic classes with well-founded contextual meaning in the 3D scenes (Table~\ref{tab:ts40k_sem_labels}).
To ensure safety and security, the land strips are further processed and partitioned into three sample types: 
\textbf{(1) Tower-radius}: Includes the environment around a power-line supporting tower, providing a comprehensive view of the surroundings relevant to the tower's location.
\textbf{(2) Power-line}: Focuses on power lines as the main actors, featuring two towers at opposite sides. This sample type offers insights into the spatial relationships and configurations of power lines and their supporting structures.
\textbf{(3) No-tower}: Represents rural terrain without supporting towers but potentially includes power lines. This sample type provides context for areas where transmission infrastructure is absent, contributing to a balanced dataset.
This categorization addresses both safety concerns and the substantial data imbalance between the transmission system and its surroundings.

\subsection{Class Distribution Analysis}

Examining the statistics of the TS40K dataset, Table~\ref{tab:ts40k_sem_labels} illustrates the distribution of various classes. Notably, the \textit{ground} and \textit{low vegetation} classes are the most prevalent, while the classes that compose the electrical transmission system, i.e., \textit{power line support tower} and \textit{power line}, are under-represented.
This imbalance is a common characteristic in datasets captured from a birds-eye view perspective and in natural environments such as rural areas. 
Other 3D benchmarks may also display class imbalance, although not to the same extent as ours. For example, SemanticKITTI~\cite{SemKITTI} shows a lower point frequency in \textit{human} classes, such as cyclist. In contrast, SensatUrban~\cite{SensatUrban}, S3DIS~\cite{armeni2017joint}, and ShapeNet~\cite{chang2015shapenet} demonstrate a balanced class distribution.
Thus, our TS40K dataset presents a new challenge for model training, as some classes naturally occur less frequently. Addressing this unbalanced distribution becomes a crucial aspect for any effective approach aiming to handle diverse real-world scenarios.

\section{Tasks and Benchmarks}

\subsection{Train/Test split statistics.}

The TS40K dataset has a total of 24,355 sample across its three sample types: 3663 in tower-radius, 3590 in power-line and 17,102 in no-tower.
For each type, 80\% of the samples were reserved for model fitting and the remaining 20\% for testing.
The splitting between training and validation is done at random at each training cycle.

\subsection{Sub-sampling Techniques.}

To address the class imbalance between the electrical transmission system and its environment, we utilize farthest point sampling (FPS)~\cite{li2022adjustable}. FPS is a commonly employed technique in 3D scene understanding applications due to its ability to preserve the geometry of 3D elements and adjust class representation. However, its drawback lies in its time complexity.
Alternative sub-sampling methods, such as inverse density importance sub-sampling (IDISS)~\cite{groh2018flex} and random point sampling (RPS)~\cite{hu2020randla}, have gained attention for their more favorable time complexities. However, IDISS does not ensure the preservation of 3D scene geometry, as it selects $K$ points with lower density within a ball of radius $l$. This not only prioritizes classes with lower density, such as noise, at the expense of others, but it may also lead to varying point densities for the same type of object across different 3D scenes, such as in lower vegetation.
On the other hand, while RPS offers time efficiency, it may eliminate many points from underrepresented classes, compromising the accuracy of object segmentation. This trade-off is unacceptable in our case, where precise detection of the power grid is paramount for ensuring its safe inspection.
As a result of employing FPS, the total density for the power grid reaches 2.9\%, representing a 1.7\% increase compared to the original data distribution shown in Table~\ref{tab:ts40k_sem_labels}.

\subsection{3D Semantic Segmentation}

\subsubsection{Task and Metrics.}
For a comprehensive evaluation of model performance in 3D semantic segmentation, we rely on the mean intersection-over-union (mIoU) metric~\cite{everingham2015pascal} across all classes:
\begin{equation}
   \text{mIoU} = \frac{1}{C} \sum_{c=1}^{C} \frac{TP_c}{TP_c + FP_c + FN_c}.
\end{equation}
Here, $TP_c$, $FP_c$, and $FN_c$ stand for true positive, false positive, and false negative predictions for class $c$, and $C$ is the total number of classes.
The mIoU serves as a key metric for assessing the segmentation accuracy of 3D point cloud models. It quantifies the degree of overlap between predicted and ground truth segmentation masks, providing a comprehensive view of the model's ability to discern distinct classes of objects within the point cloud data.


\subsubsection{Results and Discussion.}
%









\begin{table}[ht]
\caption{Benchmark results of 3D semantic segmentation baselines on the TS40K test set. We report mean IoU (mIoU \%) and per-class IoU (\%) scores.
Due to the extreme class imbalance, we showcase the results with both regular and weighted cross entropy.}
\label{tab:sem_seg}
\resizebox{\textwidth}{!}{%
\begin{tabular}{l|c|c|c|c|c|c|c}
\hline
\multicolumn{1}{c|}{Method} & Loss Function                                                                     & mIoU (\%)      & Ground         & Low Vegetation & Medium Vegetation & Power Line Support Tower & Power Line     \\ \hline
PointNet~\cite{qi2017pointnet}                    & \multirow{4}{*}{Cross Entropy}                                                    & 36.25          & 49.57          & 55.53          & 9.58              & 4.52                     & 66.73          \\
PointNet++~\cite{qi2017pointnet++}                  &                                                                                   & 40.72          & 59.05          & 55.62          & 11.42             & 2.92                     & 74.58          \\
RandLaNet~\cite{hu2020randla}                   &                                                                                   & 14.38          & 28.69          & 43.18          & 0.04              & 0                        & 0              \\
KPConv~\cite{thomas2019kpconv}                      &                                                                                   & 56.18 & 63.35 & 59.76 & 24.41    & \textbf{40.62}           &\textbf{92.75} \\ 
Point Transformer~\cite{zhao2021point,pointcept2023} & & 59.26 & 75.15 & \textbf{66.02} & 29.74 & 35.32 & 90.05 \\ 
Point Transformer V2~\cite{wu2022point,pointcept2023} & & \textbf{62.27} & \textbf{77.73} & 65.78 & \textbf{49.45} & 26.39 & 91.98 \\\hline
PointNet~\cite{qi2017pointnet}                    & \multirow{4}{*}{\begin{tabular}[c]{@{}c@{}}Weighted\\ Cross Entropy\end{tabular}} & 44.58          & 62.72          & 44.92          & 17.91              & 17.57                     & 79.79          \\
PointNet++~\cite{qi2017pointnet++}                  &                                                                                   & 46.90          & 59.03          & 55.35          & 18.57             & 21.32                     & 80.22          \\
RandLaNet~\cite{hu2020randla}                   &                                                                                   & 16.76          & 23.21          & 40.27          & 17.38              & 0.91                        & 2.02              \\
KPConv~\cite{thomas2019kpconv}                      &                                                                                   & 57.58 & 64.52 & 59.23 & 38.08    & 33.03           & 93.06 \\ 
Point Transformer~\cite{zhao2021point,pointcept2023} & & 62.67 & \textbf{77.34} & \textbf{67.90} & 32.78 & \textbf{43.80} & 91.51 \\ 
Point Transformer V2~\cite{wu2022point,pointcept2023} & & \textbf{65.58} & 77.31 & 64.22 & \textbf{48.94} & 43.42 & \textbf{93.99} \\
\hline
\end{tabular}%
}
\end{table}
Table~\ref{tab:sem_seg} showcases the benchmark results of prominent 3D semantic segmentation models on the TS40K dataset. Focusing on the first six rows, Point Transformer V2~\cite{wu2022point} achieves the highest mIoU of 62.27\%. 
Point Transformer~\cite{zhao2021point} and KPConv~\cite{thomas2019kpconv} follow with a mIoU of 59.26\% and 56.18\%, respectively, both exhibiting some struggles in segments like medium vegetation.
Interestingly, KPConv achieves the highest IoU performance in power grid segments, namely in power lines (92.75\%) and their supporting towers (40.62\%).
In contrast, RandLaNet~\cite{hu2020randla} performs the worst with a mean IoU of 14.38\%, notably failing to detect any elements of the power grid. It's essential to note that random point sampling, a characteristic of RandLaNet, wasn't a factor here, as farthest point sampling was employed in all benchmarks. 
The last six rows of Table~\ref{tab:sem_seg} show that the weighted cross entropy loss consistently improves the performance of the semantic segmentation models in the TS40K dataset.
Point Transformer V2~\cite{wu2022point} maintains its position as the top-performing baseline, achieving a mIoU of 65.58\%. 
This model excels particularly in segments such as power line (93.99\%) and medium vegetation (48.94\%). 
Similarly, Point Transformer~\cite{zhao2021point} and PointNet++~\cite{qi2017pointnet++} exhibit improvement with the weighted loss, achieving an IoU improvement in segmenting supporting towers of 8.48\% and 18.4\%, respectively.
%
Alas, these 3D semantic segmentation baselines do not exhibit a performance high enough to be safely employed in the high-risk task of power-grid maintenance. 
Industry experts recommend an IoU of over 85\% for both power lines and their supporting towers as a threshold to consider incorporating these baselines into their inspection pipelines.
This performance gap highlights the ongoing challenge of mitigating the extreme class imbalance in the TS40K dataset, emphasizing the need for research and innovation to create robust semantic segmentation models that can effectively address such imbalances.

\subsection{3D Object Detection}

\subsubsection{Task and Metrics.}
In 3D object detection, Average Precision (AP) is the metric of choice, providing a comprehensive measure of the model's effectiveness. 
It involves determining the area under the precision-recall curve and averaging across all classes:
\begin{align}
    AP_c = \int_0^1 P_c(R) \, dR \quad \text{and} \quad mAP = \frac{1}{C} \sum_{c=1}^{C} AP_c.
\end{align}
Here, $P_c(R)$ denotes the precision at recall level $R$ for class $c$. 
The integration spans the entire recall range, capturing the precision nuances. The resulting $AP_c$ offers a granular evaluation of the model's effectiveness in discerning class-specific objects within the 3D space.
This metric provides a detailed assessment of precision in 3D point cloud analysis and sheds light on the model's accuracy in classifying diverse structures within the point cloud data.

\subsubsection{Results and Discussion.}
\begin{table}[ht]
\centering
\caption{Benchmark results of 3D object detection baselines on the TS40K test set, under the 3D Average Precision (AP) metric with 11 recall points. We report mean AP and per-class AP scores.}
\label{tab:obj-det}
\begin{tabular}{l|c|c|c|c}
\hline
\multicolumn{1}{c|}{Method} & \multicolumn{1}{c|}{mAP}            & Power Line Support Tower & Power Line    & Medium Vegetation \\ \hline
SECOND~\cite{yan2018second}      & 52.68 & 32.64 & 85.09 & 40.32 \\
PointPillar~\cite{lang2019pointpillars} & 56.63 & 38.63 & 83.74 & 47.53 \\
Point-RCNN~\cite{shi2019pointrcnn}  & 57.65 & 36.54 & 88.71 & 44.26 \\
Part-$A^2$ net~\cite{shi2020points}  & 58.65 & 39.55 & 86.69 & 48.00 \\
PV-RCNN~\cite{shi2020pv}                     & \multicolumn{1}{c|}{\textbf{61.23}} & \textbf{40.32}           & \textbf{92.77} & \textbf{50.61}    \\ \hline
\end{tabular}
\end{table}
Table~\ref{tab:obj-det} presents the benchmark results garnered from the TS40K test set, showcasing the performance of leading 3D object detection baselines within the framework of OpenPCDet~\cite{openpcdet2020}.
PV-RCNN~\cite{shi2020pv} is the top-performing method with a mean AP of 61.23\%, demonstrating superior performance across all evaluated categories. Other baselines like Point-RCNN, Part-$A^2$ net, and PointPillar also exhibit competitive mean AP scores, PV-RCNN outperforms them, indicating its efficacy in 3D object detection tasks on the TS40K dataset.
Interestingly, while all methods exhibit strong performance in detecting power lines, there's a notable discrepancy in their ability to detect power line supporting towers. We attribute this phenomenon to the inherent nature of TS40K scenes, where power lines stand as isolated elements while their supporting towers are often surrounded by other objects. Moreover, the labeling process introduces complexities, as regions surrounding towers are often mislabeled due to the top-to-bottom labeling convention followed by maintenance personnel, resulting in noisy labeling, for instance the ground and noise are mislabeled as supporting towers.
Confronting these challenges is undeniably complex, given the nuances of real-life data. Nevertheless, addressing such challenges is crucial for evaluating the efficacy of 3D object detection methods.

\section{Challenges and Limitations\label{sec:challenges}}

\subsection{Noisy Labels.}\label{sec:noisy-labels}
\begin{figure}[ht]
\centering
\begin{subfigure}[b]{0.40\columnwidth}
  \includegraphics[width=\linewidth]{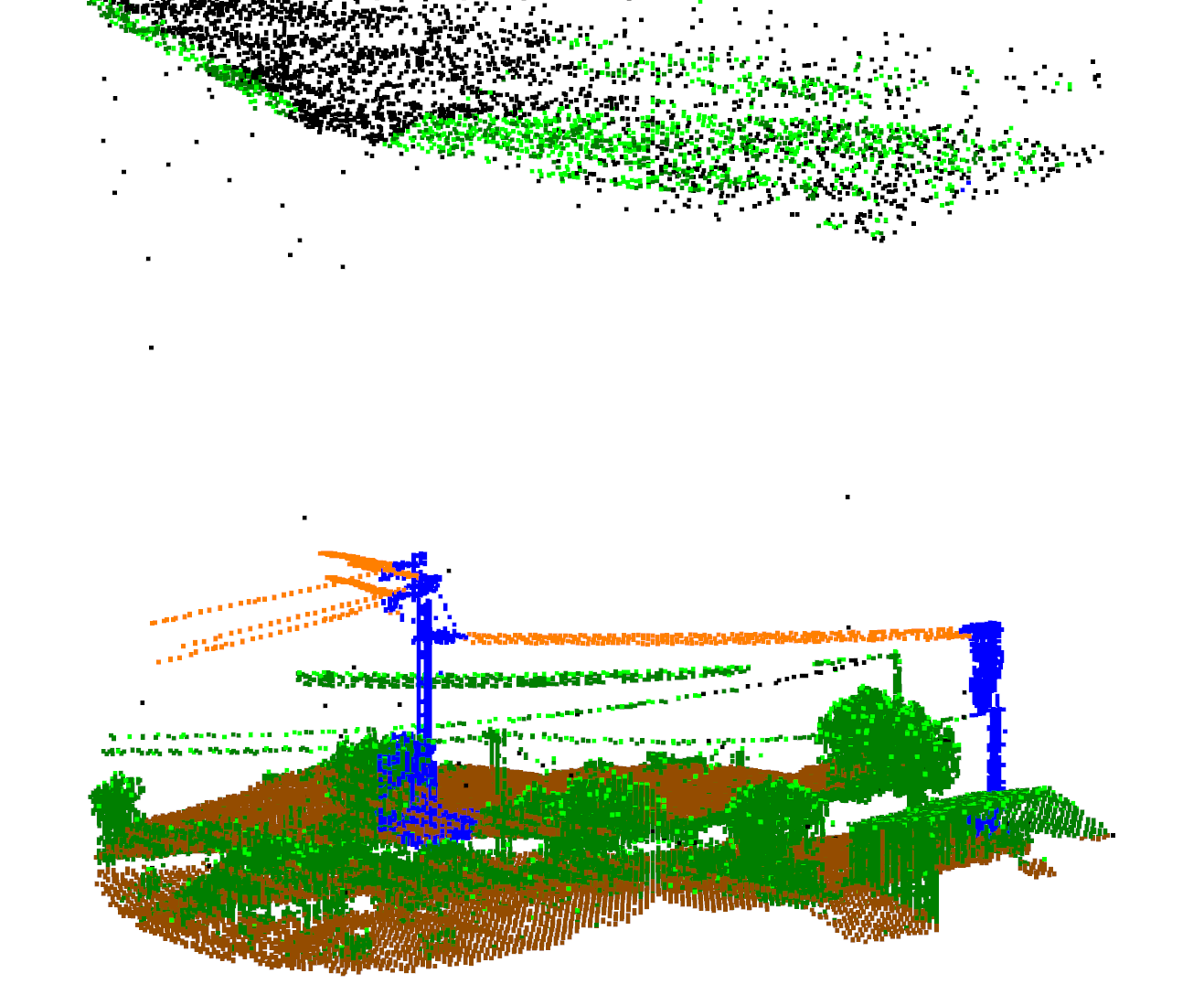}
  \caption{Noise and Power Lines mislabeled as Medium Vegetation.}
\end{subfigure}
\hfill
\begin{subfigure}[b]{0.40\columnwidth}
  \includegraphics[width=\linewidth]{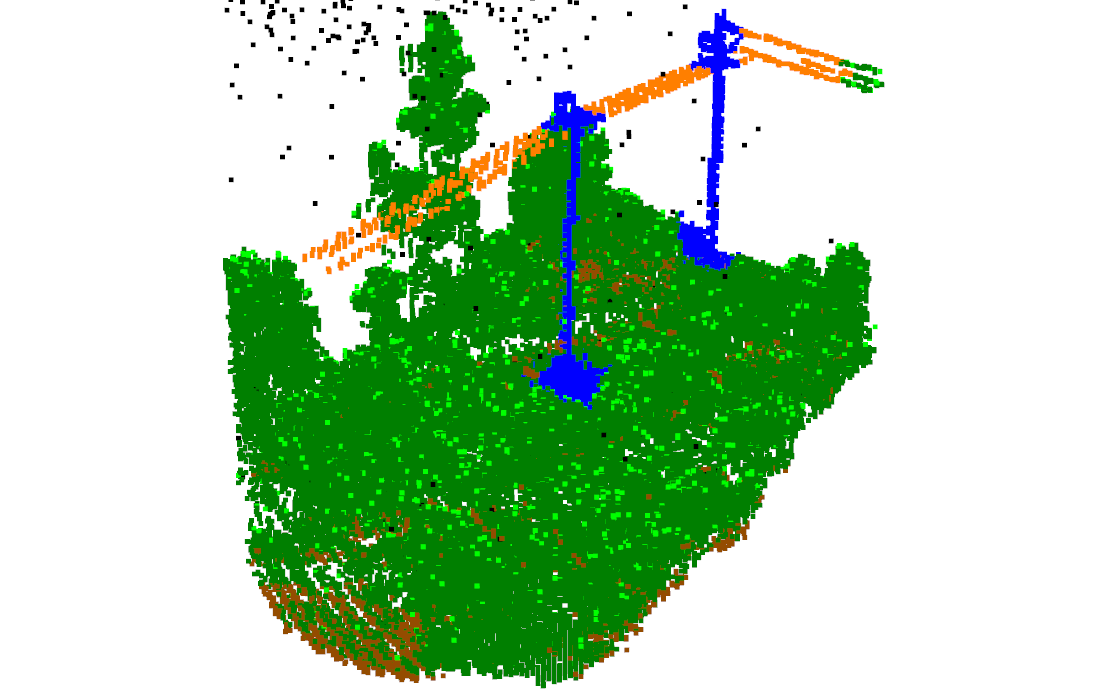}
  \caption{Ground mislabeled as Tower around supporting towers.}
\end{subfigure}

\includegraphics[width=1\columnwidth]{legend_classes.png}
\caption{In real-world 3D datasets not specifically tailored for machine learning tasks, noisy labeling can be a significant challenge. In our scenario, within the TS40K dataset, instances of mislabeled 3D elements are apparent. For instance, patches of ground might be incorrectly labeled as supporting towers, and occasional noise and power lines might be mistakenly classified as medium vegetation. This mislabeling can occur due to safety considerations around tower areas and the presence of power lines not connected to the main grid, which may not be properly identified.}
\label{fig:noisy_labels}
\end{figure}
The acquisition and labeling process of TS40K raw 3D point clouds is a long and meticulous process that focuses on safeguarding the transmission system.
The dataset's high point-density and homogeneous object density, which provide a detailed representation of rural environments, coupled with the thousands of kilometers of power grid to be closely inspected leads to labeling mishaps when mapping the maintenance classes to semantically meaningful labels. 
Specifically, some points around power line supporting towers might be misclassified due to a top-to-bottom labeling convention.
Patches of ground below towers, some power line elements and occasional noise above them are, sometimes, mistakenly labeled as part of towers. This contributes to the problem of false negative classifications.
Additionally, a number of power lines may be misclassified as medium vegetation because they are not part of the main power grid system (Fig.~\ref{fig:noisy_labels}). This could potentially result in an increased occurrence of false positive scenarios.
Thus, this mislabeling complicates 3D scene understanding evaluation.
For instance, in 3D object detection tasks, misclassified points around towers can lead algorithms to inaccurately identify tower structures, impacting performance metrics such as mAP. 
On the other hand, the inclusion of diverse scenarios ensures a broad coverage of real-world challenges. 
Such misclassification highlights the obstacles inherent to using datasets not specifically curated for machine learning tasks, emphasizing the need for more robust models that can learn to work with real-life data. 
Addressing these challenges may involve refining labeling protocols or developing algorithms robust enough to handle and adapt to the complexities of real-world data. 
This emphasizes the importance of incorporating such considerations into the development and evaluation of 3D scene understanding methodologies.

\subsection{The Impact of Extreme Class Imbalance.}
The TS40K dataset, like other real-world 3D benchmarks, presents a significant challenge in the form of extreme class imbalance, particularly prominent in power grid elements, namely power lines and their supporting towers. 
This imbalance impacts the performance of the 3D scene understanding baselines assessed on our dataset.
Overall, in both 3D semantic segmentation and 3D object detection (refer to Tables~\ref{tab:sem_seg} and~\ref{tab:obj-det}), the evaluated baselines demonstrate consistent performance trends. While power lines are efficiently detected across all models, their performance with respect to supporting towers consistently lags behind, despite having similar class densities in the dataset.
In semantic segmentation, we assess the use of weighted cross entropy to mitigate the class imbalance problem.
Point Transfomer~\cite{zhao2021point}, for instance, demonstrates a significant improvement in segmenting power lines supporting towers with the weighted loss, but it only achieves an IoU increase of 3.18\% when compared to KPConv~\cite{thomas2019kpconv} trained without the weighting scheme.  
Thus, the inability of current baselines to detect supporting towers goes beyond extreme class imbalance.
Both power lines and towers and subject to noisy labeling as described in~\ref{sec:noisy-labels}, but the discrepancy in performance cannot be justified by a small number of mislabeled points.
As such, we believe the reason for this phenomenon is twofold:
supporting towers are often densely surrounded by vegetation, while power lines are typically isolated in the air. Additionally, power lines exhibit consistent shapes, whereas supporting towers vary significantly in size, particularly in terms of height and base radius. Moreover, they feature diverse structures at the top to support power lines. These complexities contribute to the challenges faced by baseline models in achieving accurate detection of supporting towers.
Thus, our dataset presents as a great venue for the advancement of robust and reliable 3D scene understanding models in real-world scenarios.

\subsection{Is Geometry Enough?}

In our dataset, relying solely on geometric data for 3D scene understanding may not suffice. While geometric information is crucial, it might not capture all scene nuances, especially in scenarios where scene elements are closely intertwined, such as power line supporting towers in between dense vegetation.
While recent 3D benchmarks like SensatUrban~\cite{SensatUrban} incorporate RGB features alongside geometric data to capture 3D scenes comprehensively, this inclusion may not always be practical or justified in industrial settings due to sensor costs and data availability.
To address this limitation, the incorporation of RGB images could offer a promising solution. Hybrid methods~\cite{dai20183dmv,jaritz2019multi,tang2020searching,hou2022point} that fuse 2D images with 3D scenes have demonstrated improved performance in prominent benchmarks.
%
%
However, the integration of RGB information presents its own set of challenges, including data preprocessing and alignment issues. These challenges constitute pertinent areas for future research.

\section{Future Research Directions}

\paragraph{Generalization across Different Rural Environments.}

A promising avenue for future research lies in enhancing the generalization capabilities of models across diverse rural environments. 
To this end, one potential approach is to expand our dataset to encompass diverse rural settings from different geographical locations.
By incorporating scenes from various regions, our dataset would benefit from varied environmental features, such as seasonal variations and types of infrastructure, thereby enriching the dataset's diversity and realism.
Furthermore, domain adaptation techniques could further enhance the model's ability to generalize effectively across unseen environments. These techniques focus on learning invariant features that transcend specific domains, enabling models to adapt to new contexts. By leveraging domain adaptation, models trained on a heterogeneous dataset can better capture the underlying patterns present across different rural landscapes, leading to improved performance and robustness in real-world scenarios.

\section{Conclusions}
In this paper, we present the TS40K dataset, a novel contribution to the paradigm of 3D scene understanding focusing on rural electrical transmission systems in Europe. This dataset addresses a critical gap in the research landscape, offering novel challenges distinct from those encountered in urban scenarios typically explored in existing datasets. TS40K not only facilitates advancements in the crucial task of power-grid inspection but also provides unique characteristics such as high point density and absence of occlusion, which pose distinct challenges for semantic segmentation and object detection algorithms.  
Our rigorous evaluation with state-of-the-art methods showcases the dataset's significance for these tasks. Yet, challenges persist, such as a better adaptation of real data labels for learning tasks and addressing extreme class imbalance in algorithms.
We envision that TS40K will serve as a valuable resource for advancing research in rural scene understanding and infrastructure inspection. By tackling these challenges and harnessing emerging technologies, we aim to contribute to the development of safer and more robust systems for real-world applications in the future.

\section*{Acknowledgements}
\footnotesize
This research was funded by DM 351-2022 of the Italian Ministry of University and Research and with partial support from the Italian INDAM – GNAMPA group, by FCT I.P., through the strategic project NOVA LINCS (UIDB/04516/2020). It was also funded via the ENFIELD (European Lighthouse to Manifest Trustworthy and Green AI) project, European Union’s Horizon Research and Innovation Programme, Grant Agreement No. 101120657.
Views expressed are those of the author(s) only and do not reflect those of the European Union. Neither the European Union nor the granting authority can be held responsible for them.



\bibliography{biblio}
\bibliographystyle{unsrt}

\section{Appendix}

\subsection{Qualitative Performance Analysis of Point Transformer V2 on TS40K Dataset}
\begin{figure}[ht]
\centering
\begin{subfigure}[b]{0.44\columnwidth}
  \includegraphics[width=\linewidth]{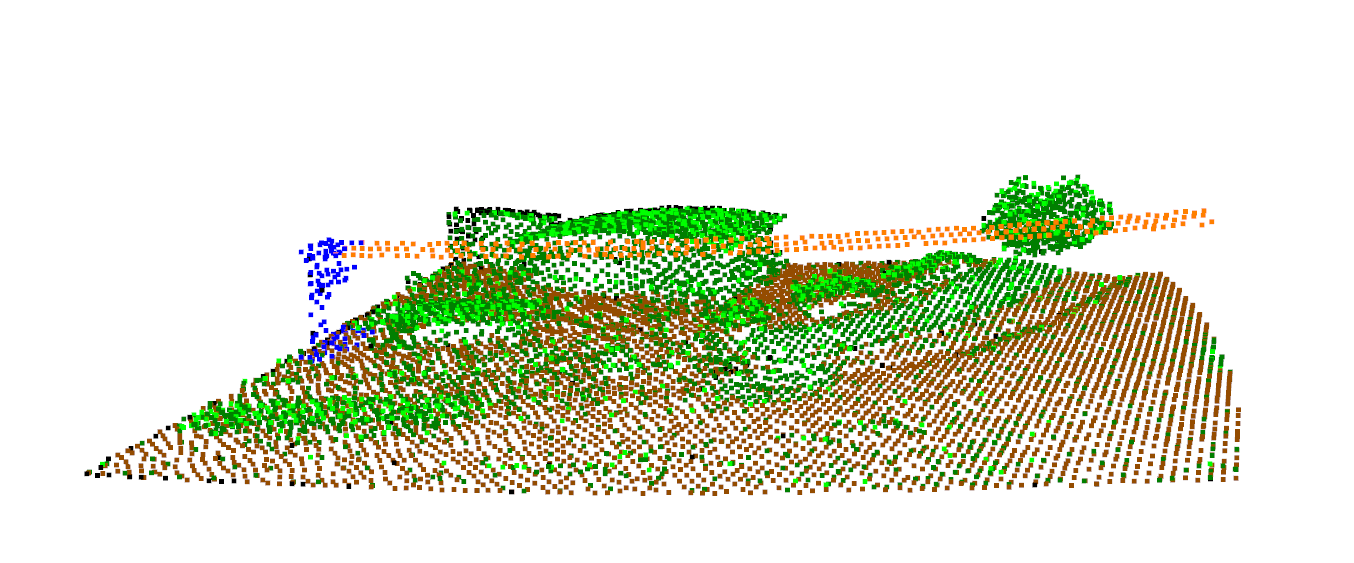}
\end{subfigure}
\hfill
\begin{subfigure}[b]{0.44\columnwidth}
  \includegraphics[width=\linewidth]{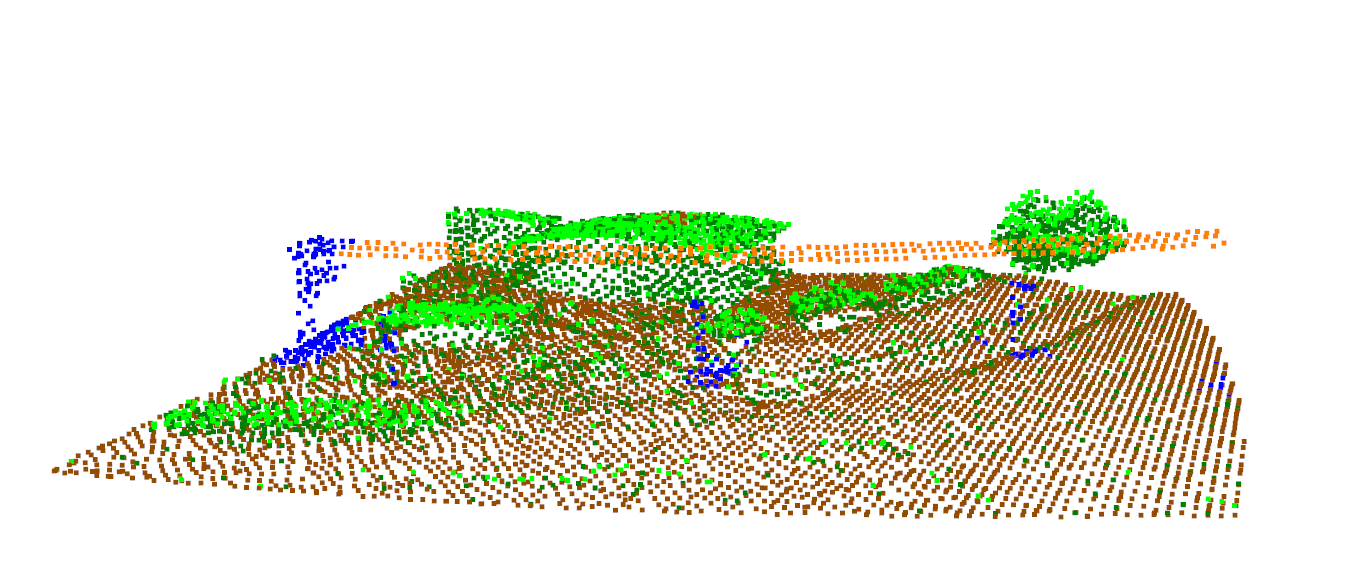}
\end{subfigure}
\\
\centering
\begin{subfigure}[b]{0.44\columnwidth}
  \includegraphics[width=\linewidth]{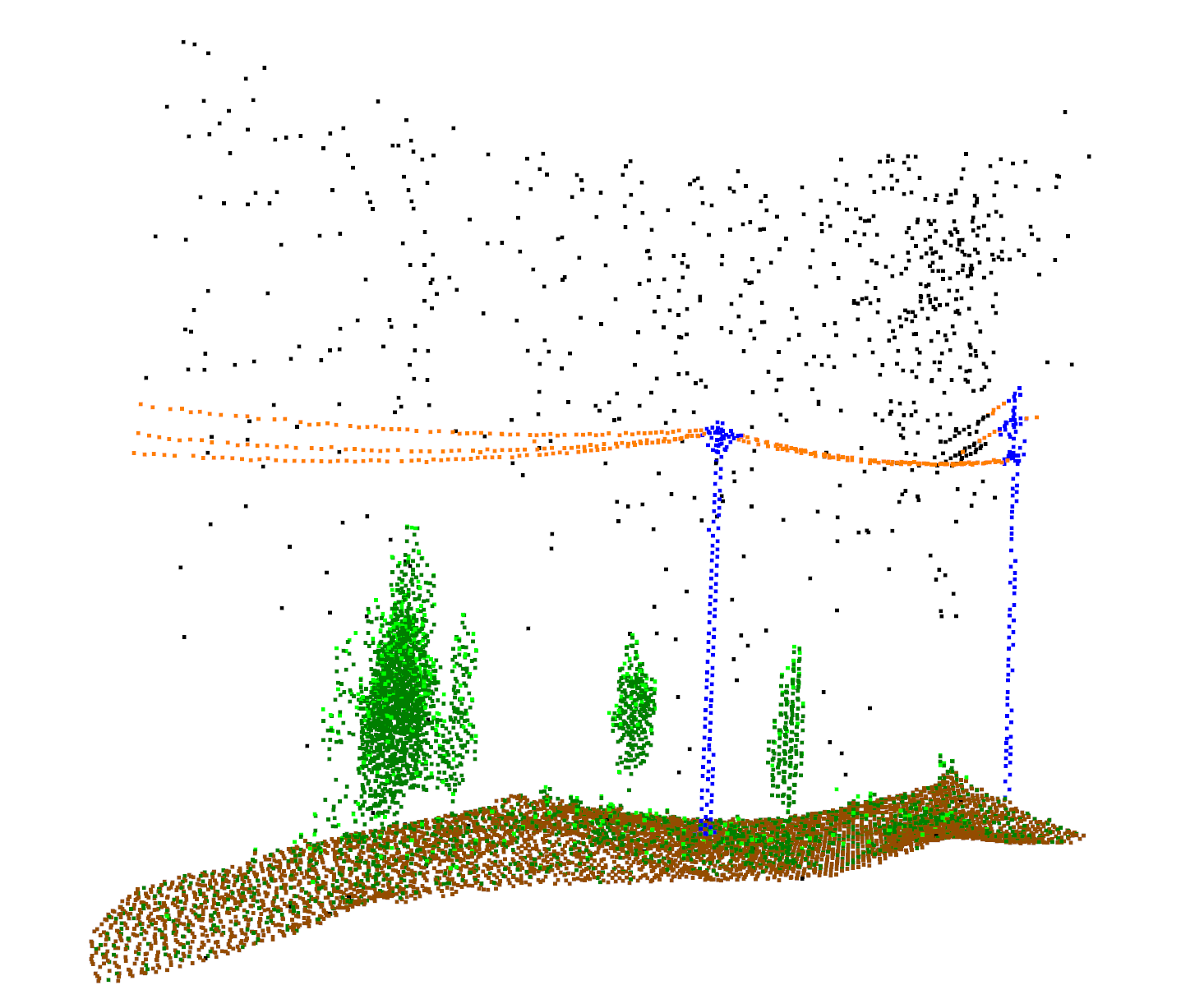}
  \caption{Ground Truth}
\end{subfigure}
\hfill
\begin{subfigure}[b]{0.44\columnwidth}
  \includegraphics[width=\linewidth]{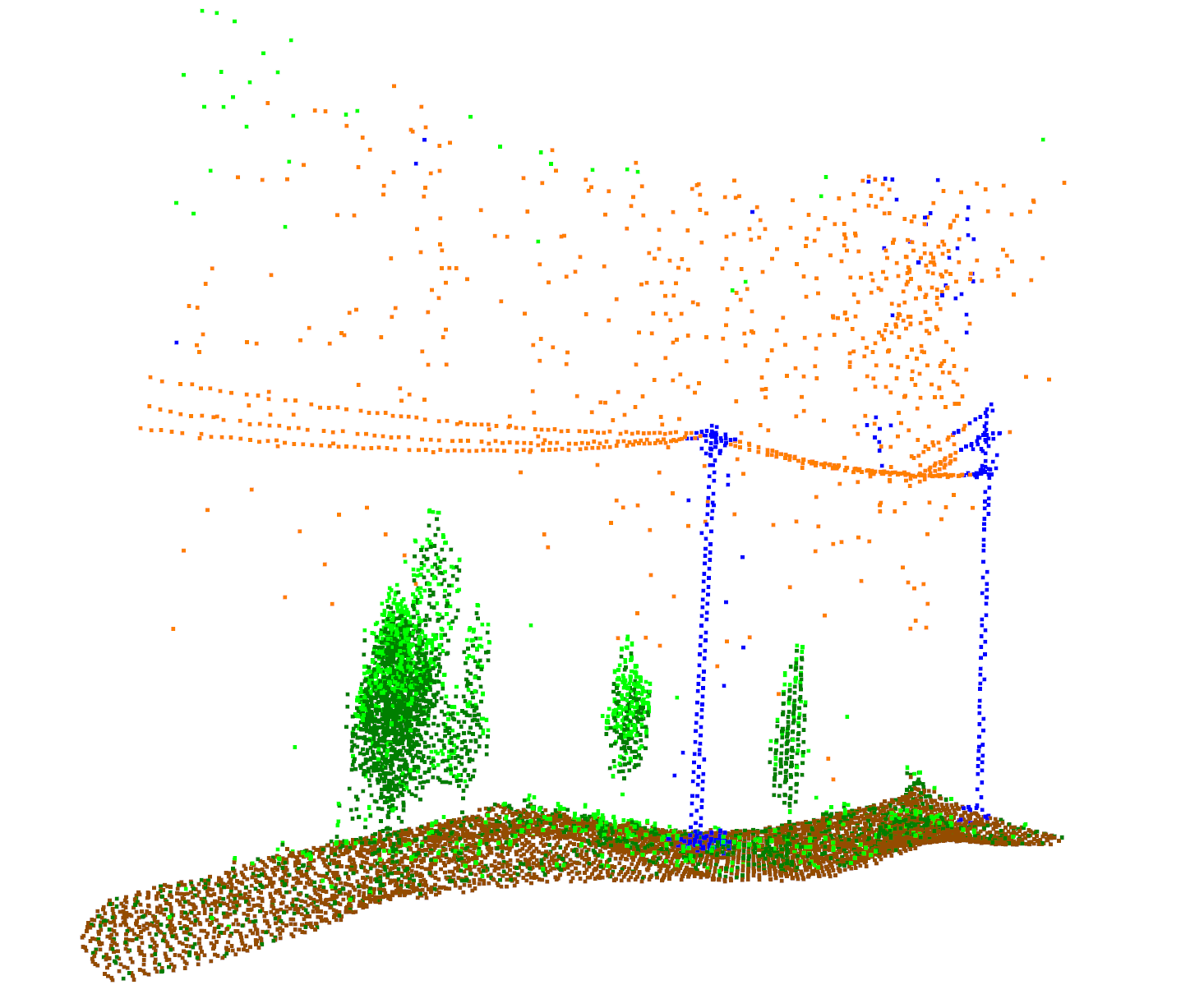}
  \caption{Point Transformer V2 Prediction}
\end{subfigure}
\\
\includegraphics[width=1\columnwidth]{legend_classes.png}
\caption{Qualitative results showcasing the performance of Point Transformer V2 (PTV2)~\cite{wu2022point} on the TS40K dataset.
In the first row, PTV2 successfully predicts the primary tower in the scene and accurately identifies smaller voltage towers, often overlooked in the ground truth annotations.
However, the second row reveals an instance where PTV2 introduces a patch of ground surrounding two towers that was absent in the original labels. This highlights the impact of noisy labels in 3D benchmarks like PTV2.}
\label{fig:ptv2-noisy-labels}
\end{figure}

\begin{figure}[ht]
\centering
\begin{subfigure}[b]{0.40\columnwidth}
  \includegraphics[width=\linewidth]{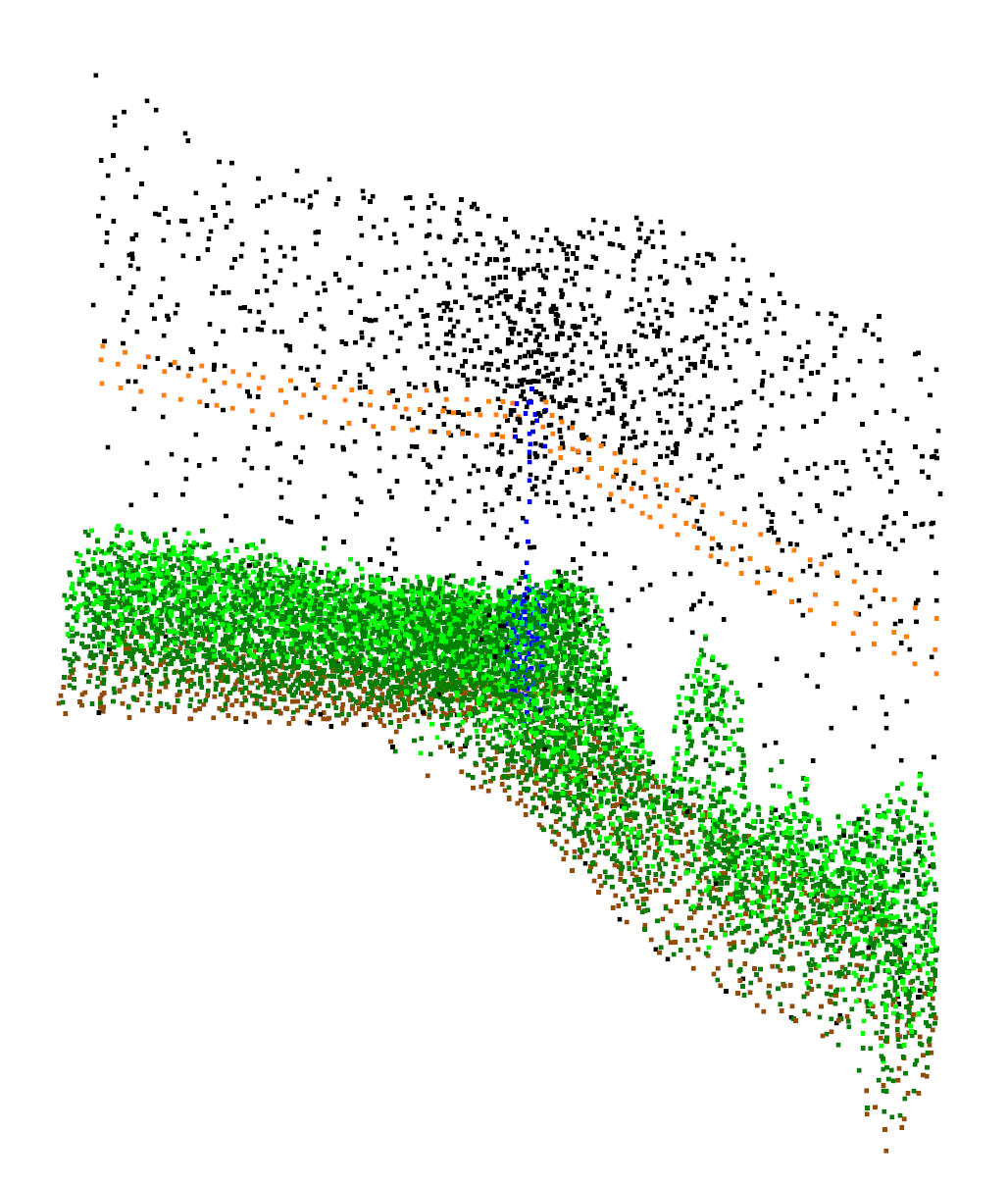}
  \caption{Ground Truth}
\end{subfigure}
\hfill
\begin{subfigure}[b]{0.40\columnwidth}
  \includegraphics[width=\linewidth]{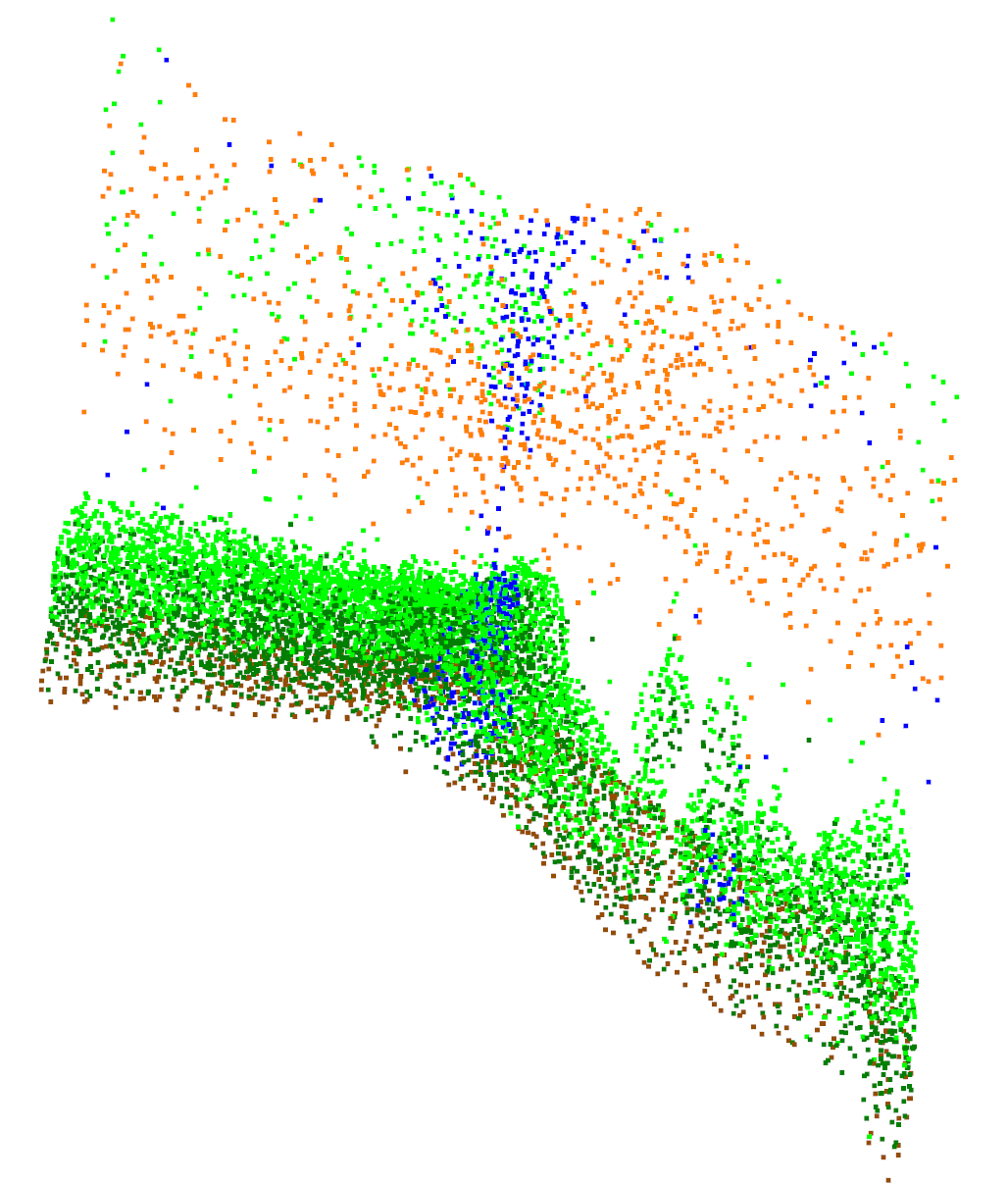}
  \caption{Point Transformer V2 Prediction}
\end{subfigure}
\\
\includegraphics[width=1\columnwidth]{legend_classes.png}
\caption{Addressing the challenge of high noise-density in Point Transformer V2 (PTV2)~\cite{wu2022point} on the TS40K dataset.
While the noise class is typically disregarded during training for 3D semantic segmentation tasks due to its lack of relevance and unpredictable distribution, certain dataset samples exhibit a high number of noise 3D points. Such a level of noise can significantly impair the accuracy of predictions in 3D benchmarks like PTV2. In this figure, the segmentation of towers and power lines becomes obscured amidst the noise, rendering the retrieval of power grid elements unattainable.
}
\label{fig:ptv2-noise}
\end{figure}

We conducted a qualitative analysis of Point Transformer V2 (PTV2)~\cite{wu2022point} on the TS40K dataset to evaluate its performance. 
Figure~\ref{fig:ptv2-noisy-labels} showcases the impact of noisy labels in the PTV2 model. In the first row, PTV2 demonstrates robustness by accurately predicting the main tower while also detecting smaller voltage towers often missed in ground truth annotations. However, our analysis reveals instances, as seen in the second row of the same Figure, where PTV2 introduces artifacts such as additional ground patches not present in the original labels. This phenomenon highlights the challenge posed by noisy labels in the TS4OK dataset, potentially affecting the reliability of model evaluations. Our findings highlight the need for further exploration into handling noisy labels in 3D scene understanding tasks.
On the other hand, Figure~\ref{fig:ptv2-noise} demonstrates the challenge of high noise-density in certain samples of the TS40K dataset.
While the noise class is typically disregarded during training for 3D semantic segmentation tasks due to its lack of relevance and unpredictable distribution, certain dataset samples exhibit a notable presence of noise 3D points. Such level of noise can severely impair the accuracy of predictions in 3D benchmarks like PTV2. As illustrated in Figure~\ref{fig:ptv2-noise}, the segmentation of towers and power lines becomes obscured amidst the noise, rendering the retrieval of power grid elements unattainable. Addressing this challenge is crucial for enhancing the reliability and robustness of 3D scene understanding models like PTV2 in real-world applications.

\subsection{Subsampling Techniques for 3D Point Clouds}
\label{sec:subsampling}
%

\begin{figure}[t]
\centering
\begin{subfigure}[b]{0.30\columnwidth}
  \includegraphics[width=\linewidth]{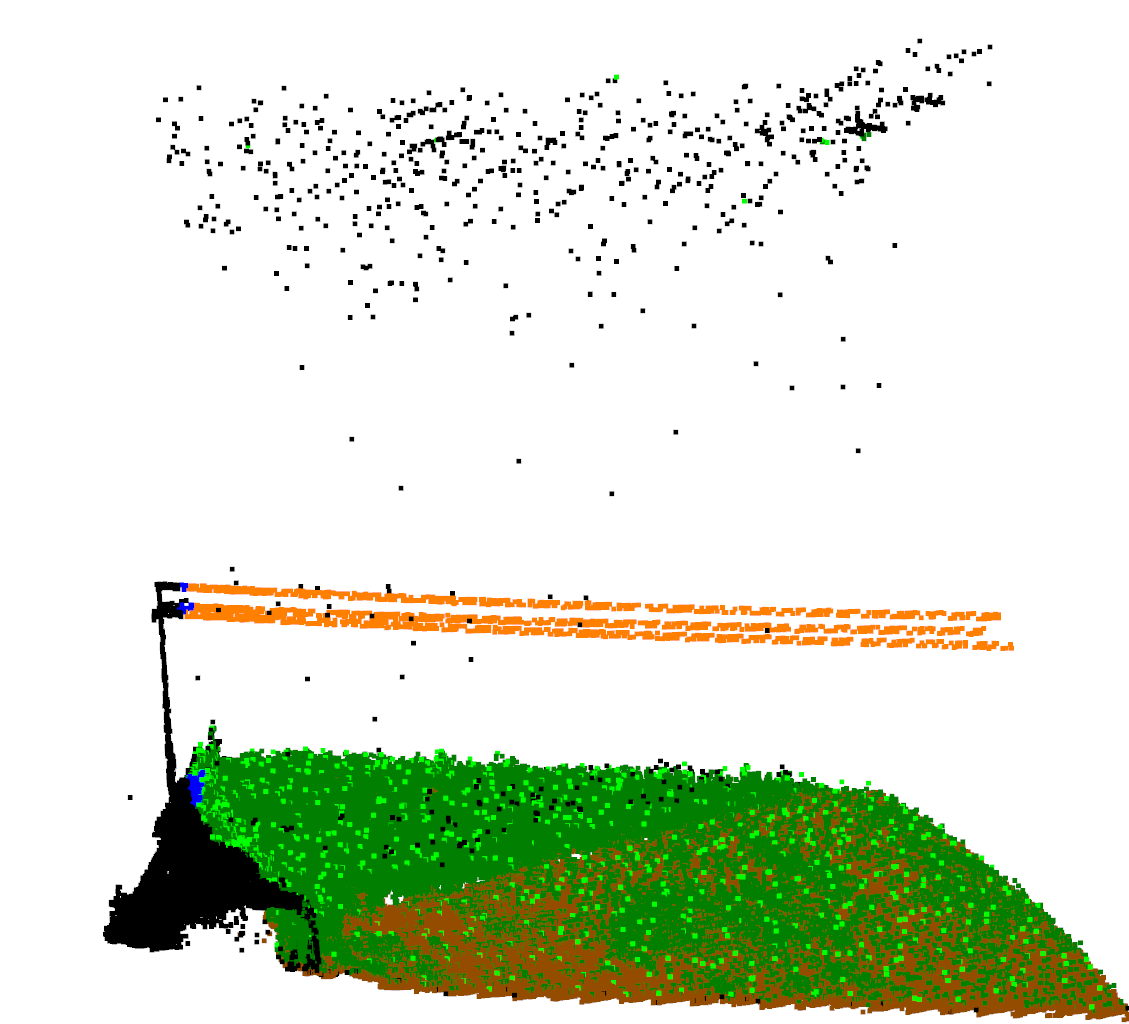}
\end{subfigure}
\hfill
\begin{subfigure}[b]{0.40\columnwidth}
  \includegraphics[width=\linewidth]{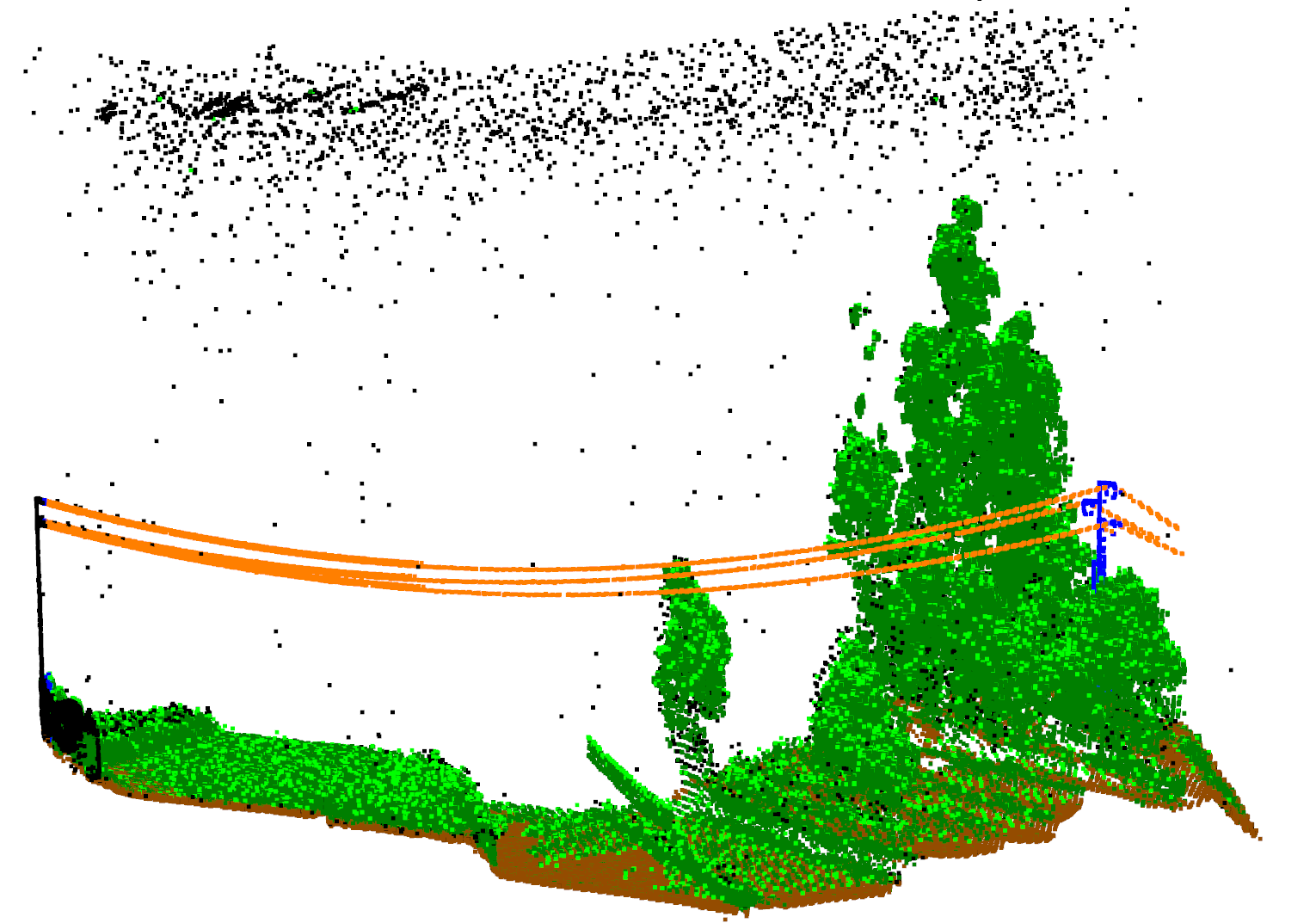}
\end{subfigure}
\hfill
\begin{subfigure}[b]{0.28\columnwidth}
  \includegraphics[width=\linewidth]{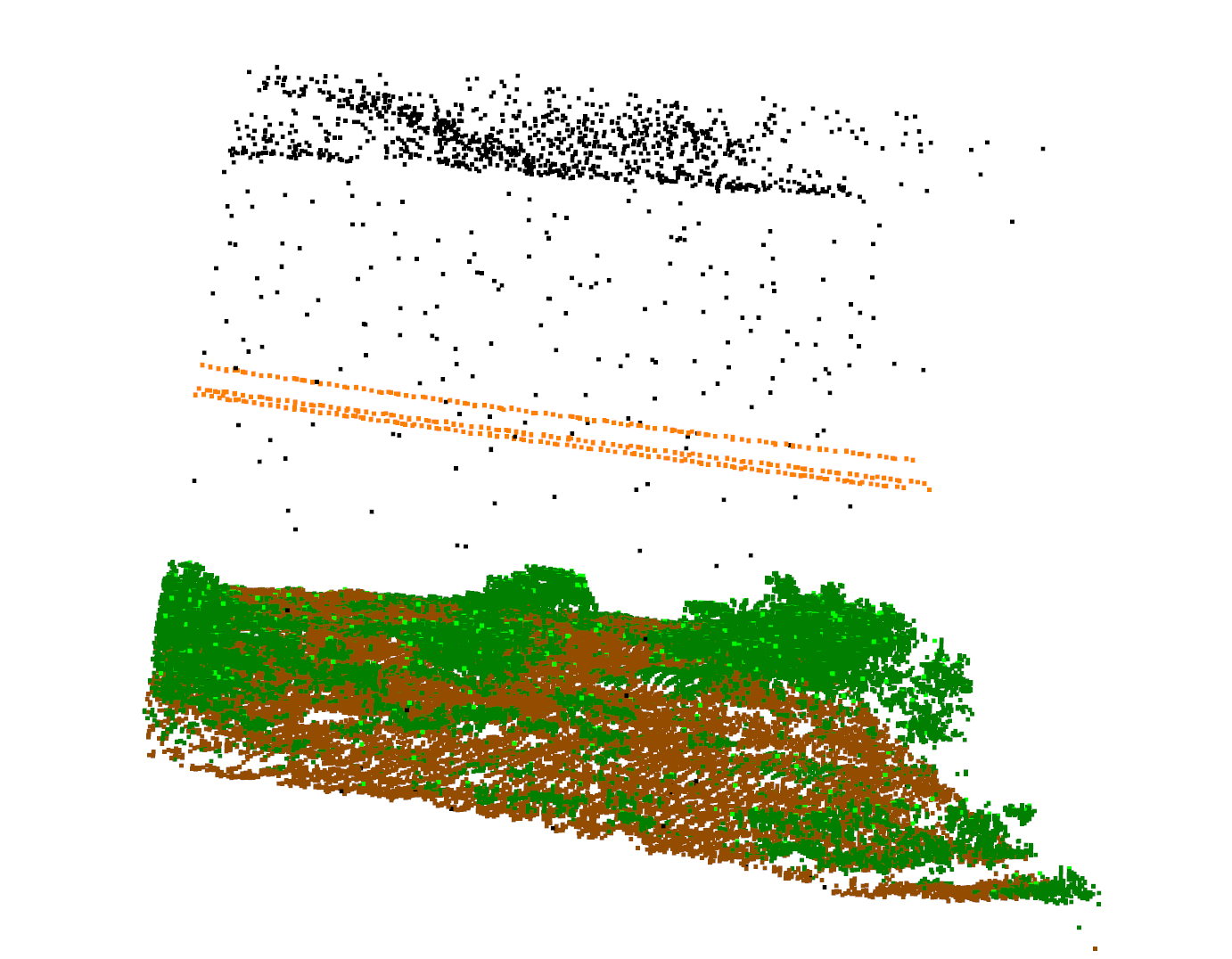}
\end{subfigure}
\\
\begin{subfigure}[b]{0.35\columnwidth}
  \includegraphics[width=\linewidth]{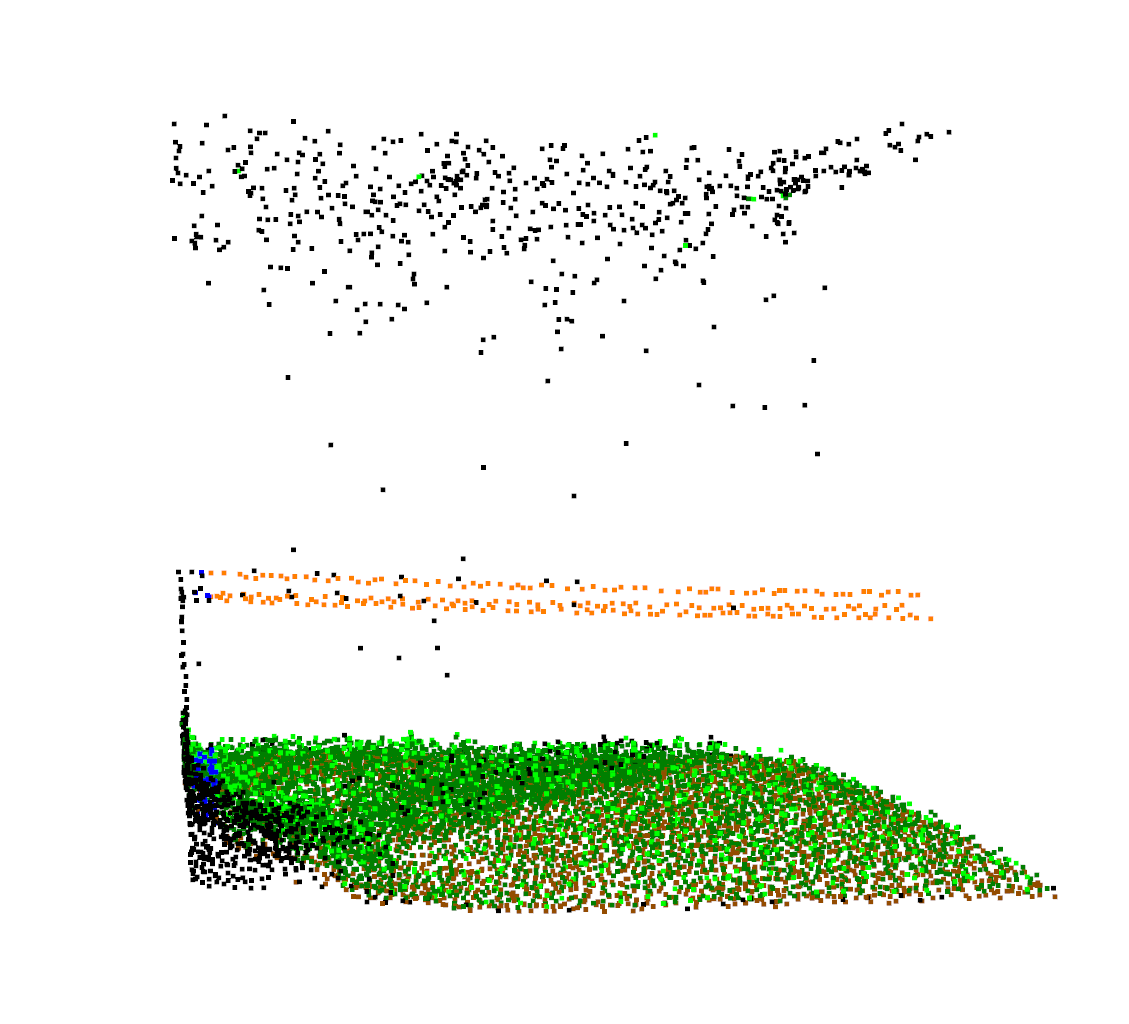}
\end{subfigure}
\hfill
\begin{subfigure}[b]{0.35\columnwidth}
  \includegraphics[width=\linewidth]{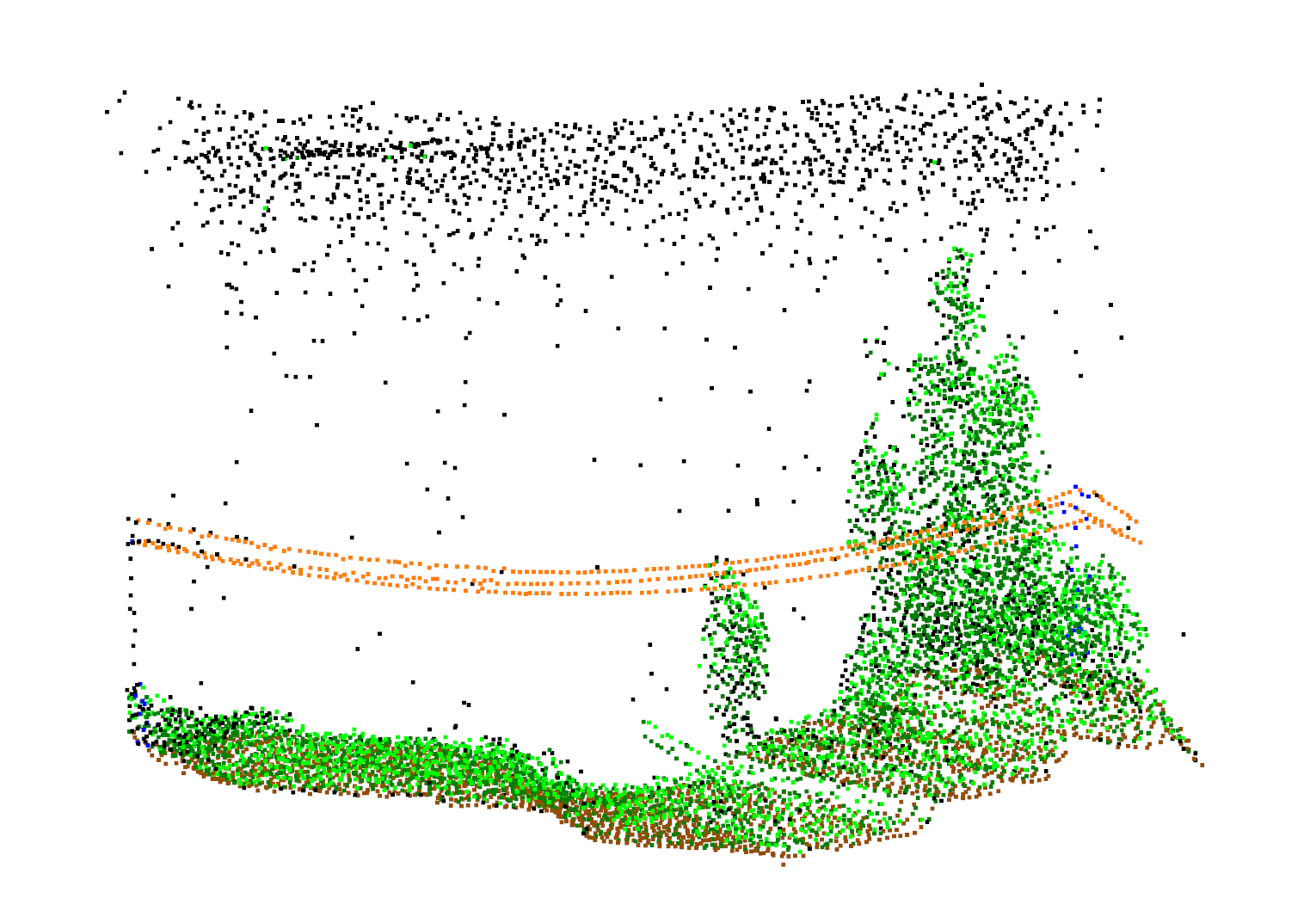}
\end{subfigure}
\hfill
\begin{subfigure}[b]{0.28\columnwidth}
  \includegraphics[width=\linewidth]{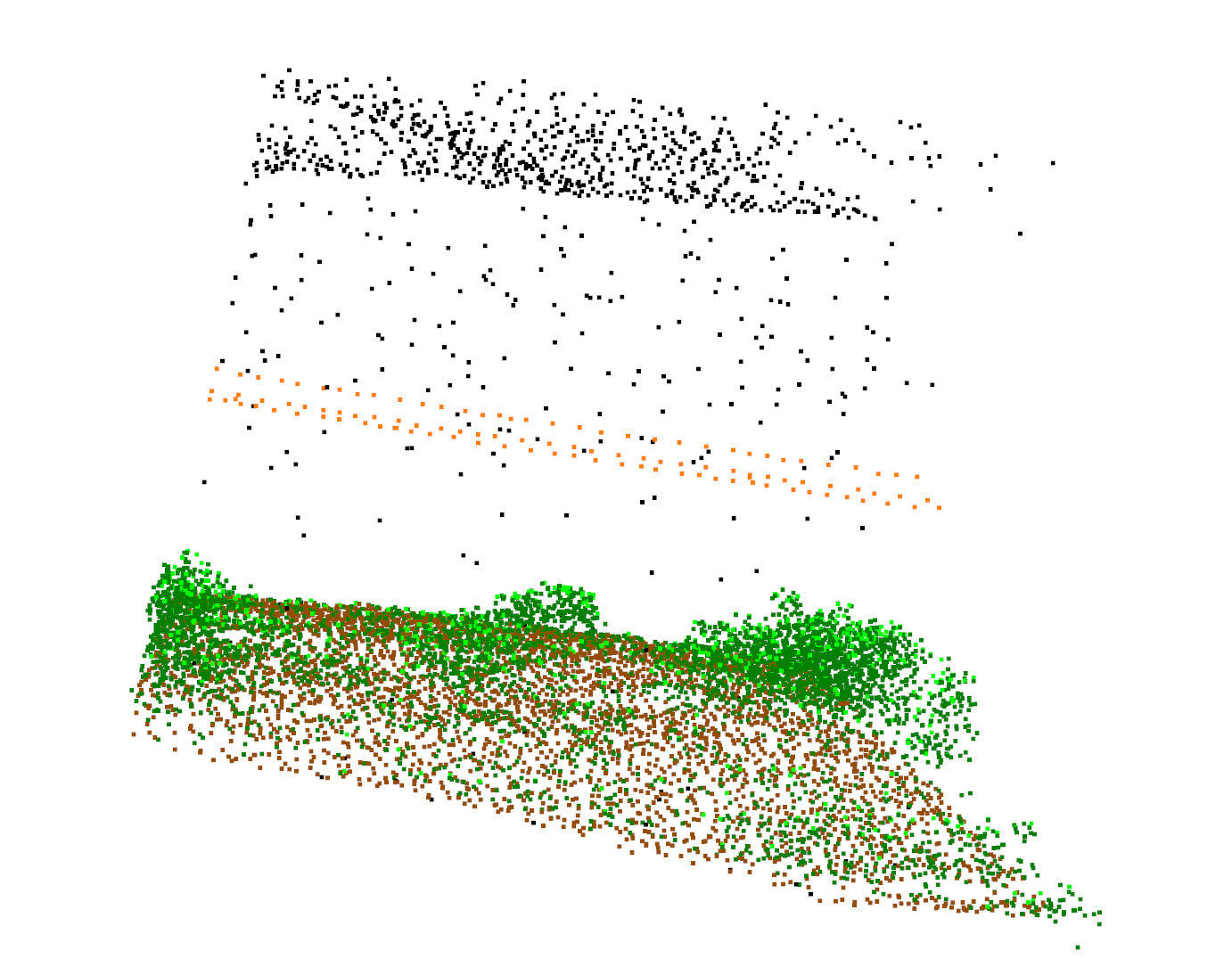}
\end{subfigure}
\centering
\\
\begin{subfigure}[b]{0.30\columnwidth}
  \includegraphics[width=\linewidth]{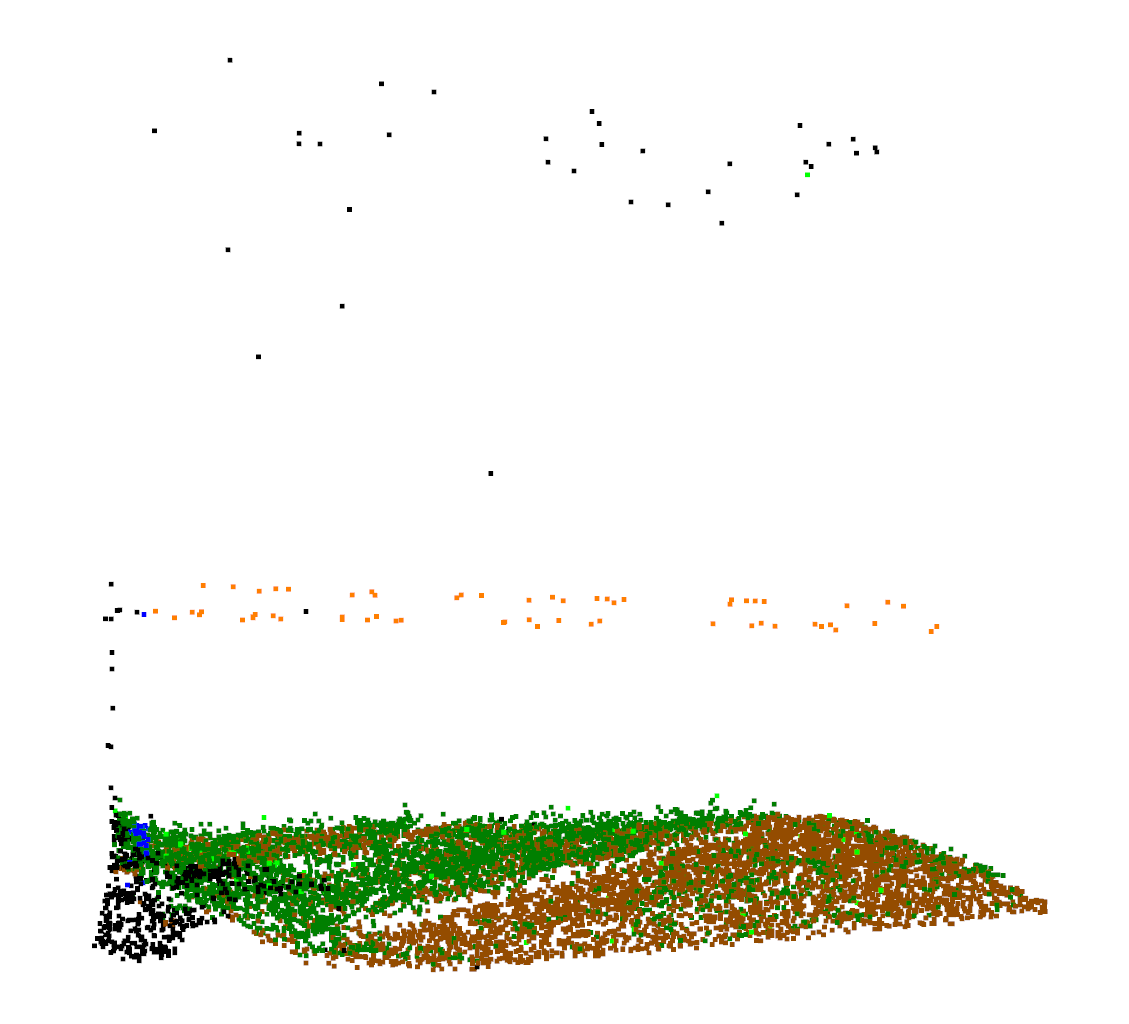}
\end{subfigure}
\hfill
\begin{subfigure}[b]{0.40\columnwidth}
  \includegraphics[width=\linewidth]{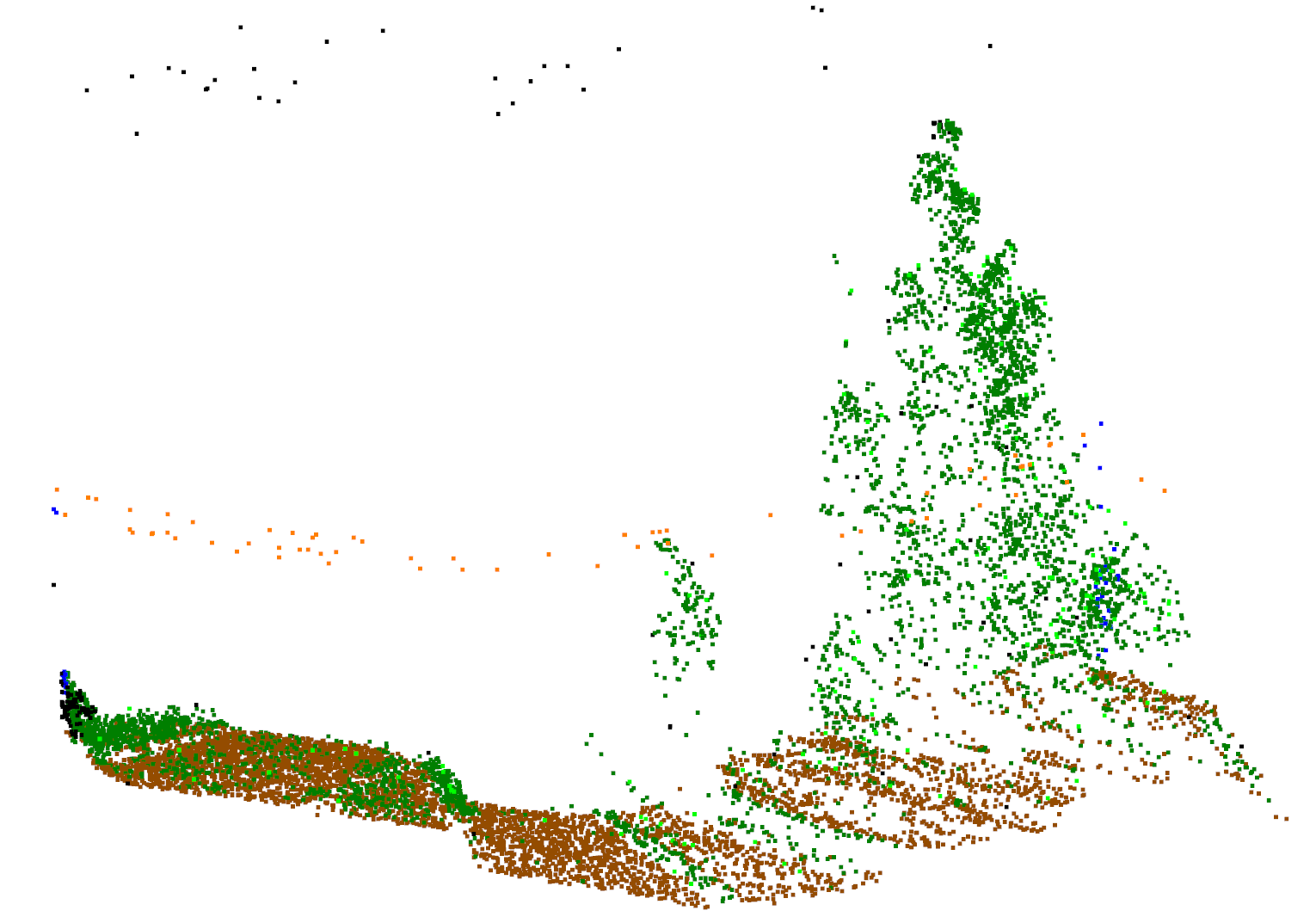}
\end{subfigure}
\hfill
\begin{subfigure}[b]{0.28\columnwidth}
  \includegraphics[width=\linewidth]{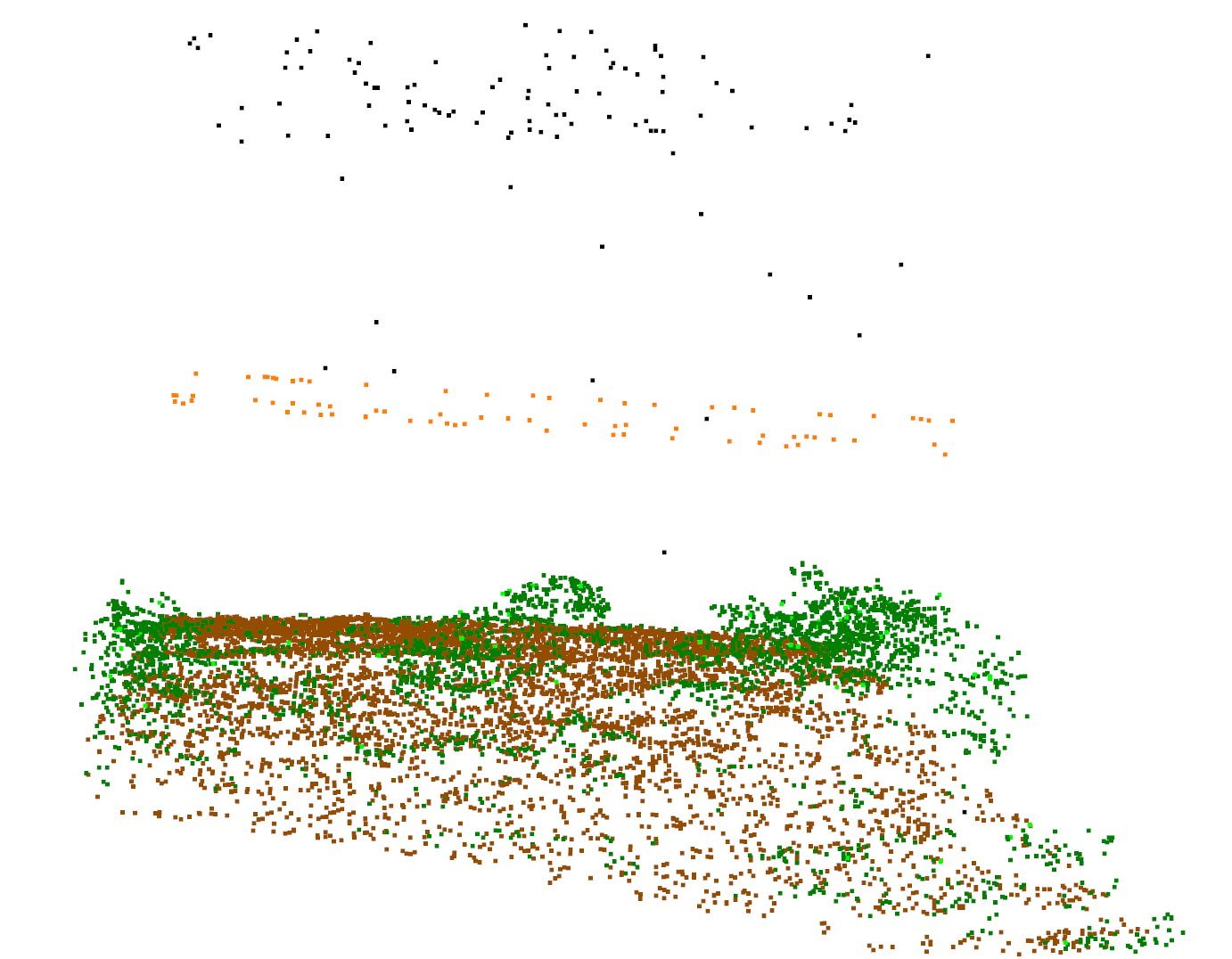}
\end{subfigure}
\\
\begin{subfigure}[b]{0.35\columnwidth}
  \includegraphics[width=\linewidth]{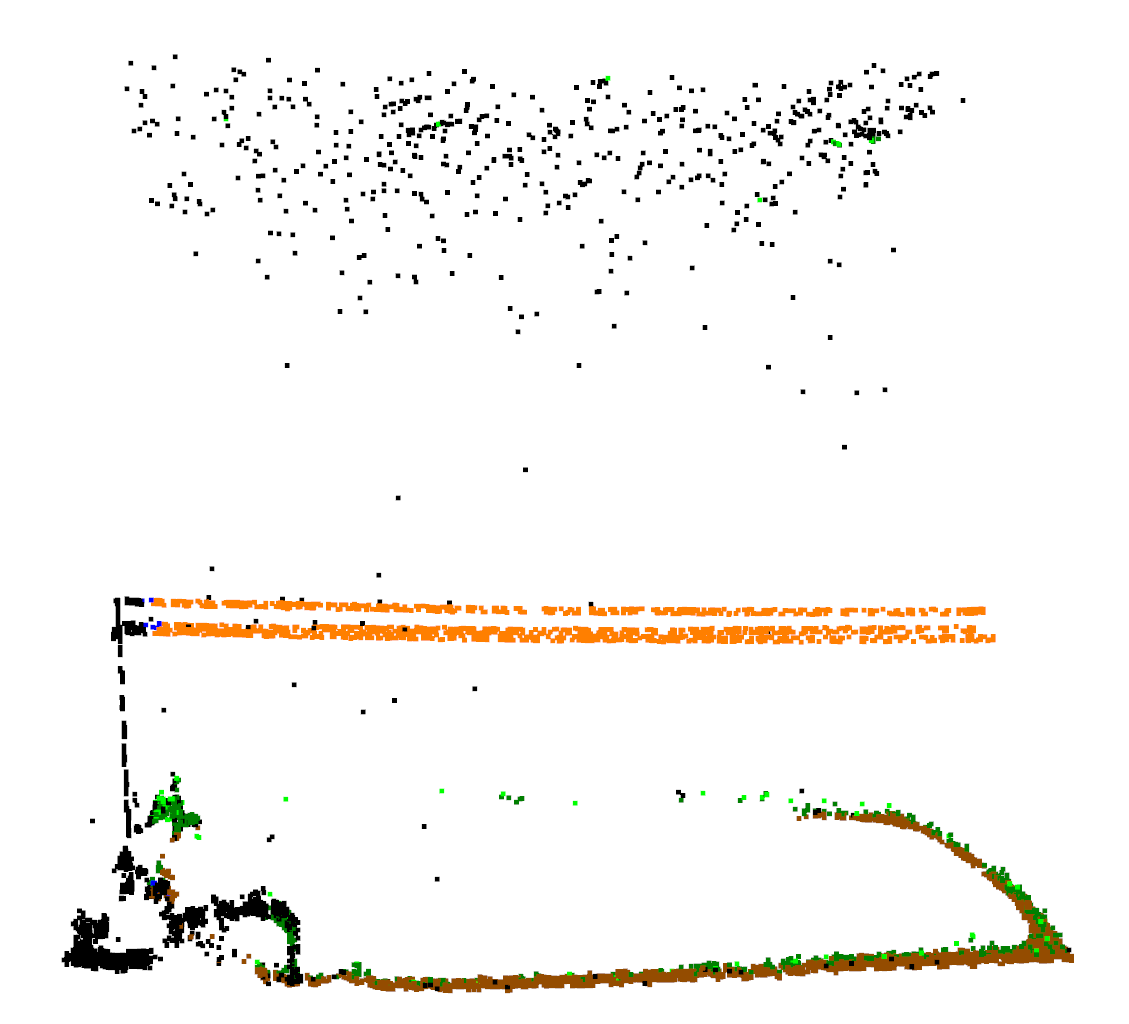}
  \caption{Tower-radius}
\end{subfigure}
\hfill
\begin{subfigure}[b]{0.35\columnwidth}
  \includegraphics[width=\linewidth]{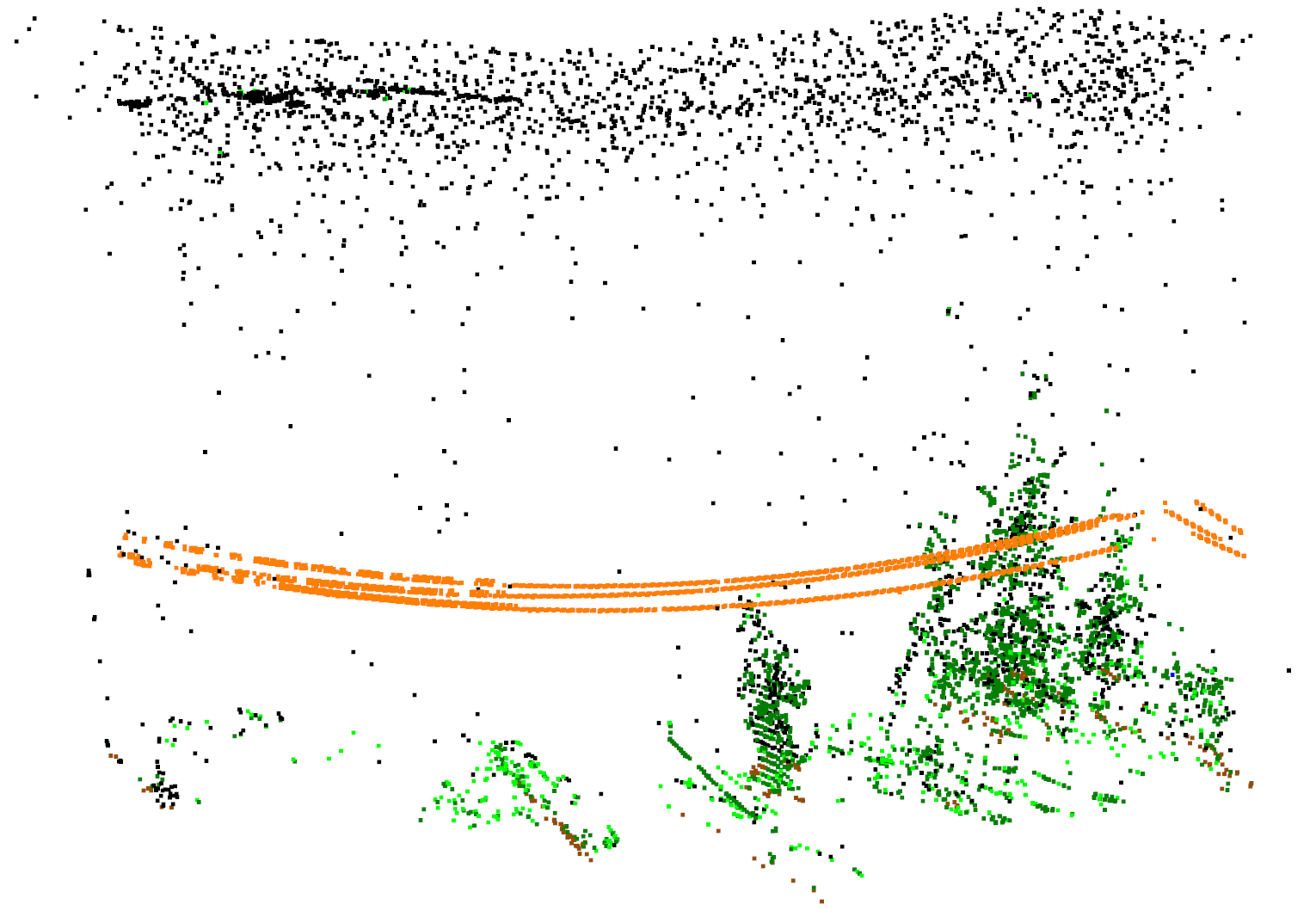}
  \caption{Power-line}
\end{subfigure}
\hfill
\begin{subfigure}[b]{0.28\columnwidth}
  \includegraphics[width=\linewidth]{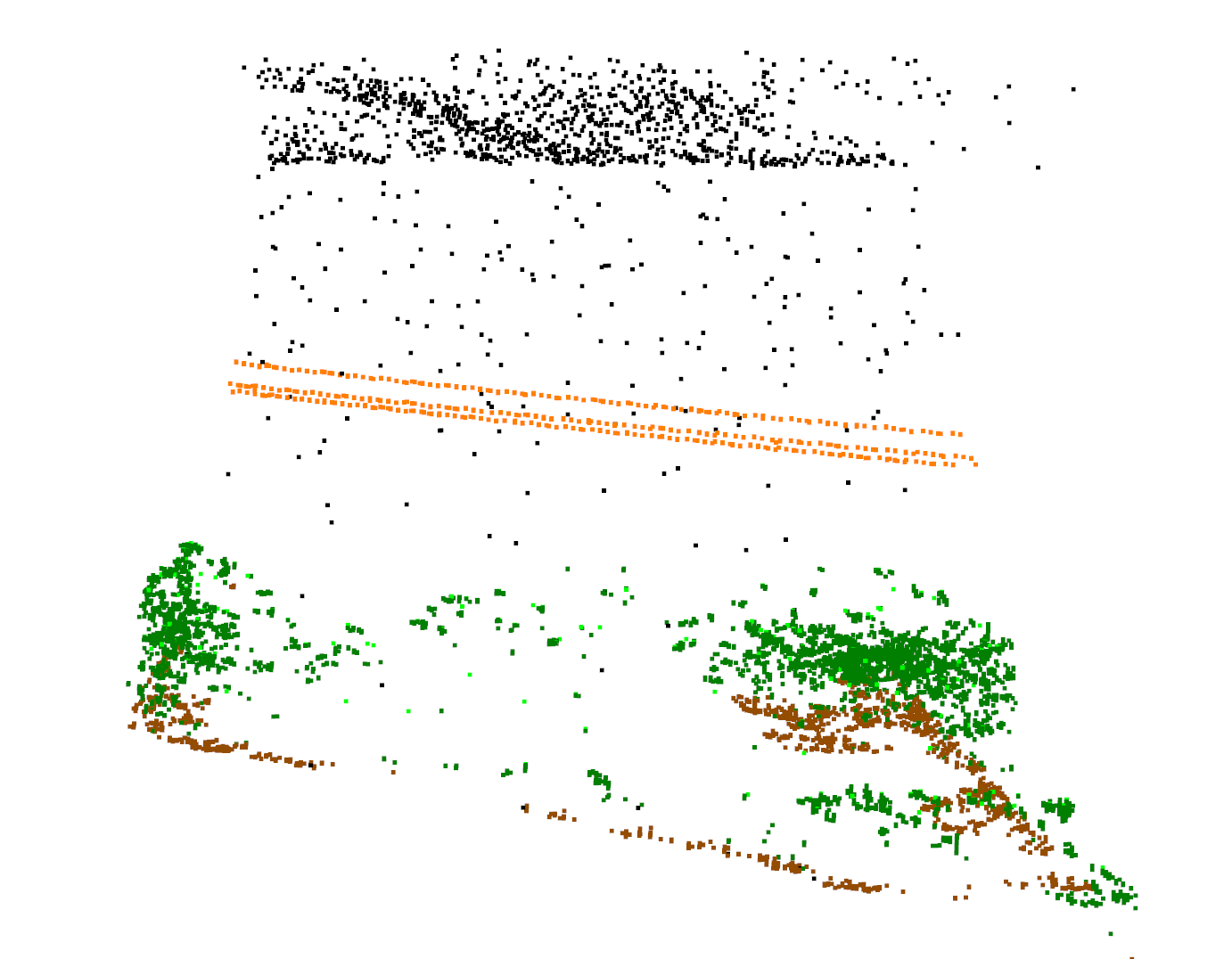}
  \caption{No-tower}
\end{subfigure}

\includegraphics[width=1\columnwidth]{legend_classes.png}
\\
\caption{ Comparison of three subsampling techniques for 3D point clouds. The top row displays the original data, while subsequent rows depict the effects of Farthest Point Sampling (FPS), Random Point Sampling (RPS), and Inverse Density Importance Subsampling (IDISS) on the dataset. FPS preserves geometry and adjusts class representation, leading to a balanced distribution with increased density for the power grid. IDISS prioritizes lower density classes, potentially compromising geometry preservation and causing inconsistent point densities, such as in the \textit{tower-radius} example where it removes ground and low vegetation. RPS, though time-efficient, may eliminate points from underrepresented classes, impacting object segmentation accuracy, as seen in the \textit{power-line} example.}
\label{fig:subsampling}
\end{figure}

\paragraph{Farthest Point Sampling.}
FPS~\cite{li2022adjustable} is a widely used technique in 3D scene understanding due to its capability to maintain geometric fidelity while balancing class representation. As illustrated in the second row of Figure~\ref{fig:subsampling}, FPS increases the density of the power grid, which is essential for accurate inspection and analysis in our application. However, it is important to note that FPS incurs a higher time complexity compared to alternative methods.

\paragraph{Random Point Sampling.}
RPS~\cite{hu2020randla} offers a time-efficient alternative to FPS but may lead to the elimination of crucial points from underrepresented classes, as shown in the third row of Figure~\ref{fig:subsampling}. In all three sample types, RPS fails to adequately preserve the representation of the power grid, consequently impacting the accuracy of object segmentation.

\paragraph{Inverse Density Importance Subsampling.}
IDISS~\cite{groh2018flex} prioritizes points with lower density, potentially compromising the preservation of geometric details. As depicted in the last row of Figure~\ref{fig:subsampling}, IDISS may lead to inconsistent point densities, particularly evident in scenarios such as \textit{tower-radius} sample, where it removes ground and low vegetation while preserving the noise of top of the 3D scene.

Overall, FPS emerges as the preferred choice for our application, striking a balance between geometric fidelity and class representation adjustment.

\end{document}